\documentclass[twocolumn,twoside,pdflatex]{IEEEtran}

\usepackage{amsmath}
\usepackage[pdftex]{color}
\usepackage[pdftex]{graphicx}
\usepackage[varg]{txfonts}
\usepackage{bm}
\usepackage{multirow}
\usepackage{url}
\usepackage{cite}
\usepackage{algorithm}
\usepackage{algorithmic}
\usepackage[right]{lineno}
\usepackage{colortbl}

\newcommand{\argmax}{\mathop{\rm arg~max}\limits}

\ifCLASSINFOpdf
\else
\fi
\newfont{\bg}{cmr9 scaled\magstep4}
\newcommand{\bigzerol}{\smash{\lower1.0ex\hbox{\bg 0}}}

\begin{document}

\title{DNN-based Source Enhancement to Increase\\ Objective Sound Quality Assessment Score}

\author{
Yuma~Koizumi$^{1}$~\IEEEmembership{Member,~IEEE,},
Kenta~Niwa$^{1}$~\IEEEmembership{Member,~IEEE,},
Yusuke~Hioka$^{2}$~\IEEEmembership{Senior Member,~IEEE},\\
Kazunori Kobayashi$^{1}$ and 
Yoichi Haneda$^{3}$~\IEEEmembership{Senior Member,~IEEE,}
\thanks{

\noindent
$^{1}$: NTT Media Intelligence Laboratories, NTT Corporation, Tokyo, Japan (e-mail: koizumi.yuma@ieee.org, {niwa.kenta, kobayashi.kazunori}@lab.ntt.co.jp)

\noindent
$^{2}$: Department of Mechanical Engineering, University of Auckland, 20 Symonds Street, Auckland, 1010 New Zealand (e-mail: yusuke.hioka@ieee.org)

\noindent
$^{3}$: Department of Informatics, The University of Electro-Communications, Tokyo, Japan (e-mail: haneda.yoichi@uec.ac.jp)

Copyright (c) 2018 IEEE. This article is the ``accepted'' version. Digital Object Identifier: 10.1109/TASLP.2018.2842156
}
}

\maketitle

\begin{abstract}
We propose a training method for deep neural network (DNN)-based source enhancement to increase objective sound quality assessment (OSQA) scores such as the perceptual evaluation of speech quality (PESQ). 
In many conventional studies, DNNs have been used as a mapping function to estimate time-frequency masks and trained to minimize an analytically tractable objective function such as the mean squared error (MSE). 
Since OSQA scores have been used widely for sound-quality evaluation, constructing DNNs to increase OSQA scores would be better than using the minimum-MSE to create high-quality output signals. However, since most OSQA scores are not analytically tractable, \textit{i.e.}, they are black boxes, the gradient of the objective function cannot be calculated by simply applying back-propagation. 
To calculate the gradient of the OSQA-based objective function, we formulated a DNN optimization scheme on the basis of \textit{black-box optimization}, which is used for training a computer that plays a game. 
For a black-box-optimization scheme, we adopt the policy gradient method for calculating the gradient on the basis of a sampling algorithm. 
To simulate output signals using the sampling algorithm, DNNs are used to estimate the probability-density function of the output signals that maximize OSQA scores. 
The OSQA scores are calculated from the simulated output signals, and the DNNs are trained to increase the probability of generating the simulated output signals that achieve high OSQA scores. 
Through several experiments, we found that OSQA scores significantly increased by applying the proposed method, even though the MSE was not minimized.
\end{abstract}

\section{INTRODUCTION}
\label{sec:intro}

\IEEEPARstart{S}{ound}-source enhancement has been studied for many years \cite{spenh,Ephraim_1984,zelinski88,Hioka2013,Niwa2016,Lightburn_2017} because of the high demand for its use for various practical applications such as automatic speech recognition \cite{Yoshioka2012,Narayanan2013,Ochiai_2017}, hands-free telecommunication \cite{Kobayashi_08,Hioka_12}, hearing aids \cite{Moore_2003,Wang_2008,Tao_2016,Zhao_2016}, and immersive audio field representation \cite{obj_base,Koizumi_2017}. 
In this study, we aimed at generating an enhanced target source with high listening quality because the processed sounds are assumed perceived by humans.

Recently, deep learning \cite{LeCun_DL_2015} has been successfully used for sound-source enhancement  
\cite{Narayanan2013, Zhao_2016,Weninger_2015,Erdogan_2015,Will_cIRM_2017,Zhao_2018,Xu_2014,Y_xu_2015_DNN,Xu_2015,Kawase2016,Gao_2016,DNN_Niwa_2017,Smagradis_2017,Chai_2017,QWang_2018,Kinoshita_2017,Arie_2016,Hershey,Pascual_2017}
. 
In many of these conventional studies, deep neural networks (DNNs) were used as a regression function to estimate 
time-frequency (T-F) masks \cite{Weninger_2015,Erdogan_2015,Will_cIRM_2017,Zhao_2018}
and/or 
amplitude-spectra of the target source \cite{Xu_2014,Y_xu_2015_DNN,Xu_2015,Kawase2016,Gao_2016,DNN_Niwa_2017,Smagradis_2017,Chai_2017,QWang_2018}.
The parameters of the DNNs were trained using back-propagation \cite{back_prop} to minimize an analytically tractable objective function such as the mean squared error (MSE) between supervised outputs and DNN outputs. 
In recent studies, advanced analytical objective functions were used such as 
the maximum-likelihood (ML) \cite{Kinoshita_2017,Chai_2017}, 
the combination of multi-types of MSE \cite{Xu_2015,Gao_2016,QWang_2018},
the Kullback-Leibler and/or Itakura-Saito divergence \cite{Arie_2016}, 
the modified short-time intelligibility measure (STOI) \cite{Zhao_2018}, 
the clustering cost \cite{Hershey}, and 
the discriminative cost of a clean target source and output signal using a generative adversarial network (GAN) \cite{Pascual_2017}.

When output sound is perceived by humans, the objective function that reflects human perception may not be analytically tractable, {\it i.e.}, it is a black-box function. 
In the past few years, objective sound quality assessment (OSQA) scores, such as the perceptual evaluation of speech quality (PESQ) \cite{PESQ} and STOI \cite{STOI}, have been commonly used to evaluate output sound quality. 
Thus, it might be better to construct DNNs to increase OSQA scores directly. 
However, since typical OSQA scores are not analytically defined ({\it i.e.}, they are black-box functions), the gradient of the objective function cannot be calculated by simply applying back-propagation.

We previously proposed a DNN training method to estimate T-F masks and increase OSQA scores \cite{Koizumi_ICASSP_2017}. 
To overcome the problem that the objective function to maximize the OSQA scores is not analytically tractable, we developed a DNN-training method on the basis of the \textit{black-box optimization} framework \cite{RL_text}, as used in predicting the winning percentage of the game Go \cite{alpha_go}. 
The basic idea of block-box optimization is estimating a gradient from randomly simulated output.
For example, in the training of a DNN for the Go-playing computer, the computer determines a ``{\it move}'' (where to put a Go-stone) depending on the DNN output. 
Then, when the computer won the game, a gradient is calculated to increase the selection probability of the selected ``{\it moves}''.
We adopt this strategy to increase the OSQA scores; some output signals are randomly simulated and a DNN is trained to increase the generation probability of the simulated output signals that achieved high OSQA scores.
For the first trial, we prepared a finite number of T-F mask templates and trained DNNs to select the best template that maximizes the OSQA score. 
Although we found that the OSQA scores increased using this method, the output performances would improve by extending the method to a more flexible T-F mask design scheme from the template-selection scheme.

\begin{figure}[ttt]
  \centering
  \includegraphics[width=85mm]{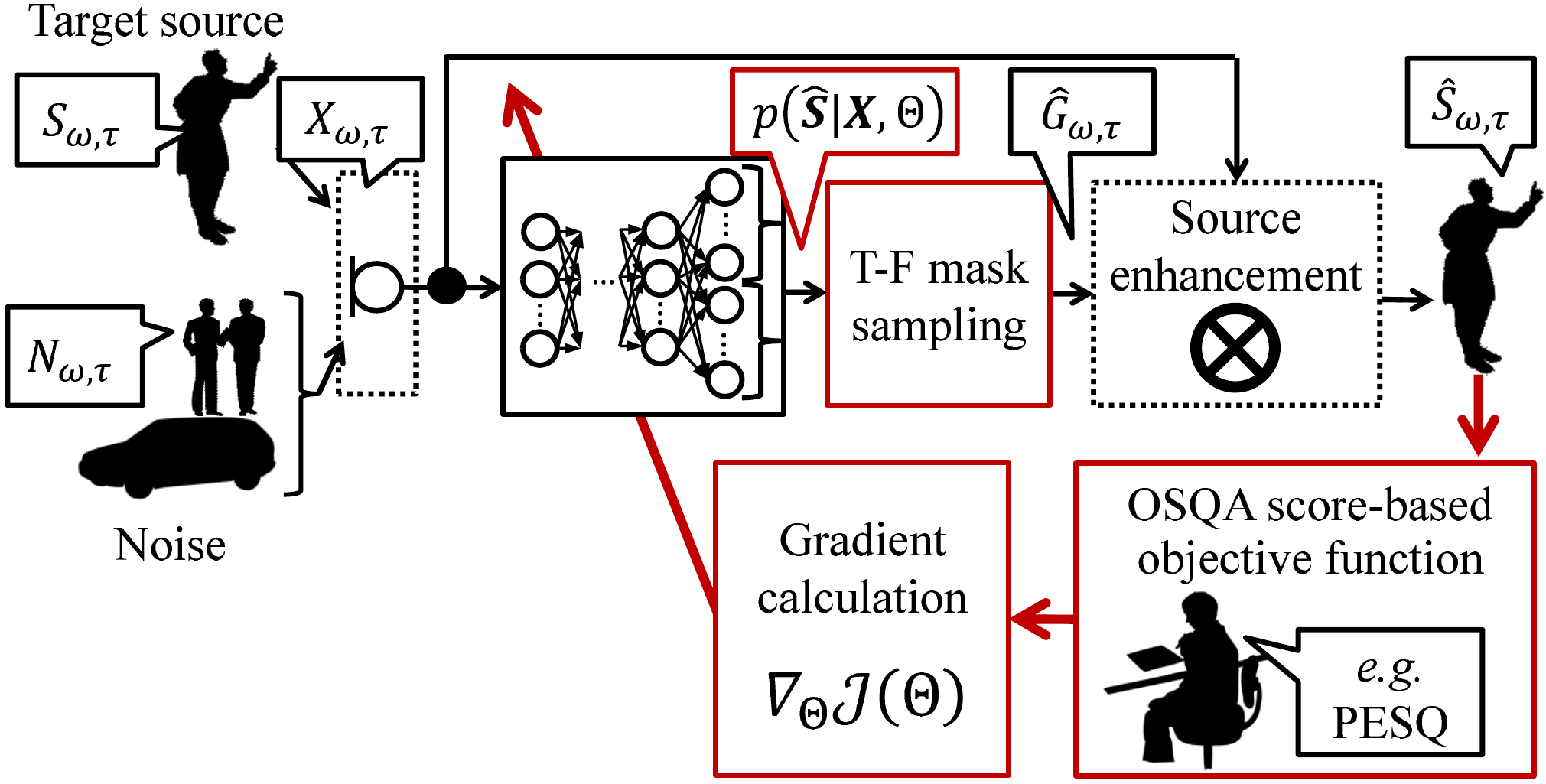}
  \caption{Concept of proposed method}
  \label{fig:concept}
\end{figure}

In this study, to arbitrarily estimate T-F masks, we modified the DNN source enhancement architecture to estimate the latent parameters in a continuous probability density function (PDF) of the T-F mask processing output signals, as shown in Fig. \ref{fig:concept}.
To calculate the gradient of the objective function, we adopt the policy gradient method \cite{Policy_Grad} as a black-box optimization scheme. 
With our method, the estimated latent parameters construct a continuous PDF as the ``{\it policy}'' of T-F-mask estimation to increase OSQA scores.
On the basis of this policy, the output signals are directly simulated using the sampling algorithm.
Then, the gradient of the DNN is estimated to increase/decrease the generation probability of output signals with high/low OSQA scores, respectively. 
The sampling from continuous PDF causes the estimate of the gradient to fluctuate, resulting in unstable training behavior. 
To avoid this problem, we additionally formulate two tricks: i) score normalization to reduce the variance in the estimated gradient, and ii) a sampling algorithm to simulate output signals to satisfy the constraint of T-F mask processing.

The rest of this paper is organized as follows. Section \ref{sec:conv_method} introduces DNN source enhancement based on the ML approach. 
In Section \ref{sec:prop_method}, we propose our DNN training method to increase OSQA scores on the basis of the black-box optimization. 
After investigating the sound quality of output signals through several experiments in Section \ref{sec:eval}, we conclude this paper in Section \ref{sec:conclusion}.


\section{CONVENTIONAL METHOD}
\label{sec:conv_method}
\subsection{Sound source enhancement with time-frequency mask}

Let us consider the problem of estimating a target source $S_{\omega, \tau} \in \mathbb{C}$, which is surrounded by ambient noise $N_{\omega, \tau} \in \mathbb{C}$. A signal observed with a single microphone $X_{\omega, \tau} \in \mathbb{C}$ is assumed to be modeled as 
\begin{equation}
X_{\omega, \tau} = S_{\omega, \tau} + N_{\omega, \tau},
\label{eq:obs}
\end{equation}
where $\omega=\{ 1,2,...,\Omega \}$ and $\tau=\{ 1,2,...,T \}$ denote the frequency and time indices, respectively.

In sound-source enhancement using T-F masks, the output signal $\hat{S}_{\omega, \tau}$ is obtained by multiplying a T-F mask by $X_{\omega, \tau}$ as
\begin{equation}
\hat{S}_{\omega, \tau} = G_{\omega, \tau} X_{\omega, \tau},
\label{eq:TF-masking}
\end{equation}
where $0 \leq G_{\omega, \tau} \leq 1$ is a T-F mask. The IRM $G_{\omega, \tau}^{\mbox{\scriptsize IRM}}$ \cite{Narayanan2013} is an implementation of T-F mask, which is defined by
\begin{equation}
G_{\omega, \tau}^{\mbox{\scriptsize IRM}} = \frac{|S_{\omega, \tau}|}{|S_{\omega, \tau}| + |N_{\omega, \tau}|}.
\label{eq:IRM}
\end{equation}
The IRM maximizes the signal-to-noise-ratio (SNR) when the phase spectrum of $S_{\omega, \tau}$ coincides with that of $N_{\omega, \tau}$. 
However, this assumption is almost never satisfied in most practical cases. To compensate for this mismatch, the phase sensitive spectrum approximation (PSA) \cite{Weninger_2015,Erdogan_2015} was proposed 
\begin{equation}
G_{\omega, \tau}^{\mbox{\scriptsize PSA}}
= \min \left( 1, \max \left( 0, \frac{|S_{\omega, \tau}|}{|X_{\omega, \tau}|} \cos \left( \theta_{\omega, \tau}^{(S)} - \theta_{\omega, \tau}^{(X)} \right) \right) \right), 
\label{eq:PSA}
\end{equation}
where $\theta_{\omega, \tau}^{(S)}$ and $\theta_{\omega, \tau}^{(X)}$ are the phase spectra of $S_{\omega, \tau}$ and $X_{\omega, \tau}$, respectively. Since the PSA $G_{\omega, \tau}^{\mbox{\scriptsize PSA}}$ is a T-F mask that minimizes the squared error between $S_{\omega, \tau}$ and $\hat{S}_{\omega, \tau}$ on the complex plane, we use this as a T-F masking scheme.

\subsection{Maximum-likelihood-based DNN training for T-F mask estimation}
\label{sec:ml_dnn}

In many conventional studies of DNN-based source enhancement, DNNs were used as a mapping function to estimate T-F masks. 
In this section, we explain DNN training based on ML estimation, on which the proposed method is based. 
Since the ML-based approach explicitly models the PDF of the target source, it becomes possible to simulate output signals by generating random numbers from the PDF.

In ML-based training, the DNNs are constructed to estimate the parameters of the conditional PDF of the target source providing the observation is given by $p(\bm{S}_{\tau} | \bm{X}_{\tau}, \Theta)$. 
Here, $\Theta$ denotes the DNN parameters. Its example on a fully connected DNN is described later (after (\ref{eq:z_return})). 
The target and observation source are assumed to be vectorized for all frequency bins as
\begin{align}
\bm{S}_{\tau} &:= (S_{1, \tau} ,...,S_{\Omega, \tau} )^{\top},\\
\bm{X}_{\tau} &:= (X_{1, \tau} ,...,X_{\Omega, \tau} )^{\top} \label{eq:vec_obs},
\end{align}
where $\top$ is transposition. 
Then $\Theta$ is trained to maximize the expectation of the log-likelihood as
\begin{align}
\Theta \gets \argmax _{\Theta} \mathcal{J}^{\mbox{\scriptsize ML}} (\Theta),
\end{align}
where the objective function $\mathcal{J}^{\mbox{\scriptsize ML}} (\Theta)$ is defined by
\begin{align}
\mathcal{J}^{\mbox{\scriptsize ML}} (\Theta) = \mathbb{E}_{\bm{S}, \bm{X}} \left[ \ln p(\bm{S} | \bm{X}, \Theta) \right],
\end{align}
and $\mathbb{E}_{x}[\cdot]$ denotes the expectation operator for $x$. 
However, since (8) is difficult to analytically calculate, the expectation calculation is replaced with the average of the training dataset as
\begin{align}
\mathcal{J}^{\mbox{\scriptsize ML}} (\Theta) \approx \frac{1}{T} \sum_{\tau=1}^{T} \ln p(\bm{S}_{\tau} | \bm{X}_{\tau}, \Theta). \label{eq:ml_obj_sum}
\end{align}
The back-propagation algorithm \cite{back_prop} is used in training $\Theta$ to maximize (\ref{eq:ml_obj_sum}). When $p(\bm{S}_{\tau} | \bm{X}_{\tau}, \Theta)$ is composed of differentiable functions with respect to $\Theta$, the gradient is calculated as
\begin{align}
\nabla_{\Theta} \mathcal{J}^{\mbox{\scriptsize ML}}(\Theta) 
&\approx \frac{1}{T} \sum_{\tau=1}^{T} 
\nabla_{\Theta}\ln p(\bm{S}_{\tau} | \bm{X}_{\tau}, \Theta ),
\label{eq:ml_obj_grad}
\end{align}
where $\nabla _x$ is a partial differential operator with respect to $x$.

\begin{figure}[ttt]
\centering
\includegraphics[width=70mm]{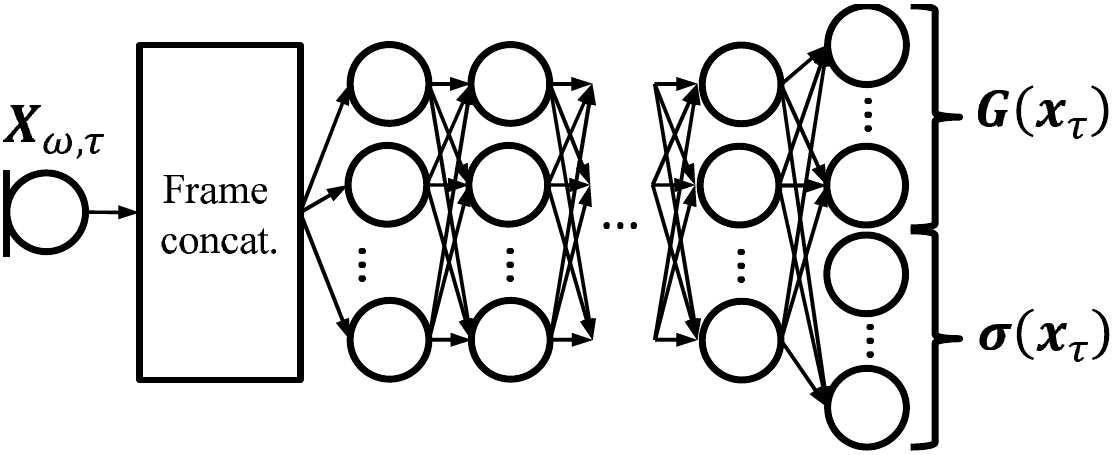}
\caption{ ML-based DNN architecture used in T-F mask estimation}
\label{fig:gauss_dnn}
\end{figure}

To calculate (\ref{eq:ml_obj_grad}), $p(\bm{S}_{\tau} | \bm{X}_{\tau}, \Theta)$ is modeled by assuming that the estimation error of $S_{\omega, \tau}$ 
 is independent for all frequency bins and follows the zero-mean complex Gaussian distribution with the variance $\sigma_{\omega, \tau}^2$.
The assumption is based on state-of-the-art methods, which train DNNs to minimize the MSE between $S_{\omega, \tau}$ and $\hat{G}_{\omega, \tau}X_{\omega, \tau}$ on the complex plane \cite{Weninger_2015,Erdogan_2015}.
The minimum-MSE (MMSE) on the complex plane is equivalent to assuming that the errors are independent for all frequency bins and follow the zero-mean complex Gaussian distribution with variance 1. 
Our assumption relaxes the assumption of the conventional methods; the variances of each frequency bin vary according to the error values to maximize the likelihood.
Thus, since $\hat{S}_{\omega, \tau}$ is given by $\hat{G}_{\omega, \tau}X_{\omega, \tau}$, $p(\bm{S}_{\tau} | \bm{X}_{\tau}, \Theta)$ is modeled by the following complex Gaussian distribution as
\begin{align}
p(\bm{S}_{\tau} | \bm{X}_{\tau}, \Theta) 
&=\prod_{\omega=1}^{\Omega} \frac{1}{ 2 \pi \sigma_{\omega, \tau}^2} 
\exp \left\{ - \frac{\left| S_{\omega, \tau} - \hat{G}_{\omega, \tau}X_{\omega, \tau} \right| ^2}{2 \sigma_{\omega, \tau}^2} \right\}. \label{eq:ML_obj}
\end{align}
In this model, it can be regarded that the MSE between $S_{\omega, \tau}$ and $\hat{S}_{\omega, \tau}$ on the complex plane is extended to the likelihood of $S_{\omega, \tau}$ defined on the complex Gaussian distribution, the mean and variance parameters of which are $\hat{S}_{\omega, \tau}$ and $\sigma_{\omega, \tau}^2$, respectively. (\ref{eq:ML_obj}) includes unknown parameters: the T-F mask $\hat{G}_{\omega, \tau}$ and error variance $\sigma_{\omega, \tau}^2$. Thus, we construct DNNs to estimate $\hat{G}_{\omega, \tau}$ and $\sigma_{\omega, \tau}^2$ from $\bm{X}_{\tau}$, as shown in Fig. \ref{fig:gauss_dnn}. The vectorized T-F masks and error variances for all frequency bins are defined as
\begin{align}
\bm{G}(\bm{x}_{\tau}) &:= \left( \hat{G}_{1, \tau} ,..., \hat{G}_{\Omega, \tau} \right)^{\top}, \label{eq:DNN_Gmean}\\
\bm{\sigma}(\bm{x}_{\tau}) &:= \left( \sigma_{1, \tau}^2 ,..., \sigma_{\Omega, \tau}^2 \right)^{\top} \label{eq:DNNvars}.
\end{align}
Here $\bm{x}_{\tau}$ is the input vector of DNNs that is prepared by concatenating several frames of observations to account for previous and future $Q$ frames as $\bm{x}_{\tau} = ( \bm{X}_{\tau-Q},..., \bm{X}_{\tau}, ..., \bm{X}_{\tau+Q} )^{\top}$, and $\bm{G}(\bm{x}_{\tau})$ and $\bm{\sigma}(\bm{x}_{\tau})$ are estimated by 
\begin{align}
\bm{G}(\bm{x}_{\tau}) 	&\gets \phi_{g} \left\{ \bm{\mathrm{W}}^{(\mu)} \bm{z}_{\tau}^{(L-1)} + \bm{\mathrm{b}}^{(\mu)} \right\}, \label{eq:DNN-mu}\\
\bm{\sigma}(\bm{x}_{\tau}) &\gets \phi_{\sigma} \left\{ \bm{\mathrm{W}}^{(\sigma)} \bm{z}_{\tau}^{(L-1)} + \bm{\mathrm{b}}^{(\sigma)} \right\} + C_{\sigma}, \label{eq:DNN-sigma}\\
\bm{z}_{\tau}^{(l)} 		&= \phi_{h} \left\{ \bm{\mathrm{W}}^{(l)} \bm{z}_{\tau}^{(l-1)} + \bm{\mathrm{b}}^{(l)} \right\}, \label{eq:z_return}
\end{align}
where $C_{\sigma}$ is a small positive constant value to prevent the variance from being very small. Here, $l$, $L$, $\bm{\mathrm{W}}^{(l)}$, and $\bm{\mathrm{b}}^{(\cdot)}$ are the layer index, number of layers, weight matrix, and bias vector, respectively.
$\bm{\mathrm{W}}^{(\mu)}, \bm{\mathrm{W}}^{(\sigma)}$ are the weight matrices and $\bm{\mathrm{b}}^{(\mu)}, \bm{\mathrm{b}}^{(\sigma)}$ are the bias vectors to estimate the T-F mask and variance, respectively.
The DNN parameters are composed of $\Theta = \{ \bm{\mathrm{W}}^{(\mu)}, \bm{\mathrm{b}}^{(\mu)}, \bm{\mathrm{W}}^{(\sigma)}, \bm{\mathrm{b}}^{(\sigma)}, \bm{\mathrm{W}}^{(l)}, \bm{\mathrm{b}}^{(l)} | l \in(2,...,L-1) \}$. 
The functions $\phi_{g}$, $\phi_{\sigma}$, and $\phi_{h}$ are nonlinear activation functions, and in conventional studies, sigmoid and exponential functions were used as an implementation of $\phi_{g}$ \cite{Weninger_2015,Erdogan_2015} and $\phi_{\sigma}$ \cite{Kinoshita_2017}, respectively. 
The input vector $\bm{x}_{\tau}$ is passed to the first layer of the network as $\bm{z}_{\tau}^{(1)} = \bm{x}_{\tau}$.

\section{PROPOSED METHOD}
\label{sec:prop_method}

Our proposed DNN-training method increases OSQA scores. With the proposed method, the policy gradient method \cite{Policy_Grad} is used to statistically calculate the gradient with respect to $\Theta$ by using a sampling algorithm, even though the objective function is not differentiable. 
However, sampling-based gradient estimation would frequently make the DNN training behavior become unstable. 
To avoid this problem, we introduce two tricks: i) score normalization that reduces the variance in the estimated gradient (in Sec. \ref{sec:prop:B}), and ii) a sampling algorithm to simulate output signals to satisfy the constraint of T-F mask processing (in Sec. \ref{sec:prop:C}). Finally, the overall training procedure of the proposed method is summarized in Sec. \ref{sec:procedure}.

\subsection{Policy gradient-based DNN training for T-F mask estimation}
\label{sec:prop:A}

Let $\mathcal{B}(\hat{\bm{S}},\bm{X})$ be a scoring function that quantifies the sound quality of the estimated sound signal $\hat{\bm{S}}:= (\hat{S}_{1} ,...,\hat{S}_{\Omega} )^{\top}$ defined by (\ref{eq:TF-masking}).
To implement $\mathcal{B}(\hat{\bm{S}},\bm{X})$, subjective evaluation is simple.
However, it would be difficult to use in practical implementation because DNN training requires a massive amount of listening-test results.
Thus, $\mathcal{B}(\hat{\bm{S}},\bm{X})$ quantifies the sound quality based on OSQA scores, as shown in Fig. \ref{fig:concept}, and the details of its implementation are discussed in Sec. \ref{sec:prop:B}.
We assume $\mathcal{B}(\hat{\bm{S}},\bm{X})$ is non-differentiable with respect to $\Theta$, because most OSQA scores are black-box functions.

Let us consider the expectation maximization of $\mathcal{B}(\hat{\bm{S}},\bm{X})$ as a metric of performance of the sound-source enhancement that increases OSQA scores as
\begin{align}
\mathbb{E}_{\hat{\bm{S}}, \bm{X}} \left[ \mathcal{B}(\hat{\bm{S}},\bm{X}) \right]
= \iint \mathcal{B}(\hat{\bm{S}},\bm{X}) p(\hat{\bm{S}}, \bm{X} ) d\hat{\bm{S}} d \bm{X}. \label{eq:exp_B}
\end{align}
Since the output signal $\hat{\bm{S}}$ is calculated from the observation $\bm{X}$, we decompose the joint PDF $p(\hat{\bm{S}}, \bm{X} )$ into the conditional PDF of the output signal given the observation $p(\hat{\bm{S}}| \bm{X} )$ and the marginal PDF of the observation $ p( \bm{X} )$ as $p(\hat{\bm{S}}, \bm{X} ) = p(\hat{\bm{S}}| \bm{X} ) p( \bm{X} )$. Then, (\ref{eq:exp_B}) can be reformed as
\begin{align}
\mathbb{E}_{\hat{\bm{S}}, \bm{X}} \left[ \mathcal{B}(\hat{\bm{S}},\bm{X}) \right]
= \int p(\bm{X}) \int \mathcal{B}(\hat{\bm{S}},\bm{X}) p(\hat{\bm{S}} | \bm{X} ) d\hat{\bm{S}} d \bm{X}. \label{eq:J_raw_base}
\end{align}
We use DNNs to estimate the parameters of the conditional PDF of the output signal $p(\hat{\bm{S}} | \bm{X}, \Theta )$, as with the case of ML-based training. For example, the complex Gaussian distribution in (\ref{eq:ML_obj}) can be used as $p(\hat{\bm{S}} | \bm{X}, \Theta )$. To train $\Theta$, $\mathbb{E}_{\hat{\bm{S}}, \bm{X}} [ \mathcal{B}(\hat{\bm{S}},\bm{X})]$ is used as an objective function by replacing the conditional PDF $p(\hat{\bm{S}} | \bm{X} )$ with $p(\hat{\bm{S}} | \bm{X}, \Theta )$ as
\begin{align}
\mathcal{J}(\Theta)	
&= \mathbb{E}_{\hat{\bm{S}}, \bm{X}} \left[ \mathcal{B}(\hat{\bm{S}},\bm{X}) \right], \label{eq:J_Eform}\\
&= \int p(\bm{X}) \int \mathcal{B}(\hat{\bm{S}},\bm{X}) p(\hat{\bm{S}} | \bm{X}, \Theta ) d\hat{\bm{S}} d \bm{X} . \label{eq:J_base}
\end{align}
Since $\mathcal{B}(\hat{\bm{S}},\bm{X})$ is non-differentiable with respect to $\Theta$, the gradient of (\ref{eq:J_base}) cannot be analytically obtained by simply applying back-propagation. Hence, we apply the policy-gradient method \cite{Policy_Grad}, which can statistically calculate the gradient of a black-box objective function. 
By assuming that the function form of $\mathcal{B}(\hat{\bm{S}},\bm{X})$ is smooth, $\mathcal{B}(\hat{\bm{S}},\bm{X})$ is a continuous function and its derivative exists.
In addition, we assume $p(\hat{\bm{S}} | \bm{X}, \Theta)$ is composed with differentiable functions with respect to $\Theta$.
Then, the gradient of (\ref{eq:J_base}) can be calculated using a log-derivative trick \cite{Policy_Grad} $\nabla_x p(\bm{x}) = p(\bm{x}) \nabla_x \ln p(\bm{x})$ as
\begin{align}
\nabla _{\Theta} \mathcal{J}(\Theta)	
&= \int p(\bm{X}) \int \mathcal{B}(\hat{\bm{S}},\bm{X}) \nabla _{\Theta} p(\hat{\bm{S}} | \bm{X}, \Theta ) d\hat{\bm{S}} d \bm{X}, \label{eq:dJ_base}\\
&= \mathbb{E}_{\bm{X}} \left[ \mathbb{E}_{\hat{\bm{S}}| \bm{X}} \left[ 
\mathcal{B}(\hat{\bm{S}},\bm{X}) \nabla _{\Theta} \ln p(\hat{\bm{S}} | \bm{X}, \Theta ) \right] \right] \label{eq:PG_exp}.
\end{align}
Since the expectation in (\ref{eq:PG_exp}) cannot be analytically calculated, the expectation with respect to $\bm{X}$ is approximated by averaging the training data, and the average of $\hat{\bm{S}}$ is calculated using the sampling algorithm as
\begin{align}
\nabla _{\Theta} \mathcal{J}(\Theta)
&\approx 
 \frac{1}{T} \sum_{\tau=1}^{T} \frac{1}{K} \sum_{k=1}^{K} 
\mathcal{B}(\hat{\bm{S}}_{\tau}^{(k)},\bm{X}_{\tau}) 
\nabla _{\Theta} \ln p(\hat{\bm{S}}_{\tau}^{(k)} | \bm{X}_{\tau}, \Theta ), \label{eq:prop_obj_grad}\\
\hat{\bm{S}}_{\tau}^{(k)} &\sim p(\hat{\bm{S}} | \bm{X}_{\tau}, \Theta ) \label{eq:monte},
\end{align}
where $\hat{\bm{S}}_{\tau}^{(k)}$ is the $k$-th simulated output signal and $K$ is the number of samplings, which is assumed to be sufficiently large. The superscript $(k)$ represents the variable of the $k$-th sampling, and $\sim$ is a sampling operator from the right-side distribution. The details of the sampling process for (\ref{eq:monte}) are described in Sec. \ref{sec:prop:C}.

Most OSQA scores, such as PESQ, are designed for their scores to be calculated using several time frames such as one utterance of a speech sentence. Since $\mathcal{B}(\hat{\bm{S}}_{\tau}^{(k)},\bm{X}_{\tau})$ of every time frame $\tau$ cannot be obtained, the gradient cannot be calculated by (\ref{eq:prop_obj_grad}). Thus, instead of using the average of $\tau$, we use the average of $\mathcal{I}$ utterances. We define the observation of the $i$-th utterance as $\bm{\mathrm{X}}^{(i)} := ( \bm{X}^{(i)}_{1}, ..., \bm{X}^{(i)}_{T^{(i)}} )$, and the $k$-th output signal of the $i$-th utterance as $\hat{\bm{\mathrm{S}}}^{(i,k)} := ( \hat{\bm{S}}^{(i,k)}_{1}, ..., \hat{\bm{S}}^{(i,k)}_{T^{(i)}} )$. Then the gradient can be calculated as
\begin{align}
\nabla _{\Theta} \mathcal{J}(\Theta) &\approx \frac{1}{\mathcal{I}} \sum_{i=1}^{\mathcal{I}} \nabla _{\Theta} \mathcal{J}^{(i)}(\Theta),
\label{eq:total_obj}\\
\nabla _{\Theta} \mathcal{J}^{(i)}(\Theta) &\approx \sum_{k=1}^{K}
\frac{\mathcal{B} \left( \hat{\bm{\mathrm{S}}}^{(i, k)}, \bm{\mathrm{X}}^{(i)} \right)}{KT^{(i)}} 
\sum_{\tau=1}^{T^{(i)}} \nabla _{\Theta} \ln p(\hat{\bm{S}}_{\tau}^{(i, k)} | \bm{X}_{\tau}^{(i)}, \Theta ), 
\label{eq:mean_obj}
\end{align}
where $T^{(i)}$ is the frame length of the $i$-th utterance, and we assume that the output signal of each time frame is calculated independently. The details of the deviation of (\ref{eq:total_obj}) are described in the Appendix \ref{sec:app_A}.

\subsection{Scoring-function design for stable training}
\label{sec:prop:B}

We now introduce a design of a scoring function $\mathcal{B}(\hat{\bm{\mathrm{S}}},\bm{\mathrm{X}})$ to stabilize the training process. 
Because the expectation for the gradient calculation in (\ref{eq:PG_exp}) is approximated using the sampling algorithm, the training may become unstable. 
One reason for unstable training behavior is that the variance in the estimated gradient becomes large in accordance with the large variance in the scoring-function output \cite{Policy_Grad}.
To stabilize the training, instead of directly using a raw OSQA score as $\mathcal{B}(\hat{\bm{\mathrm{S}}},\bm{\mathrm{X}})$, a normalized OSQA score is used to reduce its variance. 
Hereafter, a raw OSQA score calculated from $\bm{\mathrm{S}}$, $\bm{\mathrm{X}}$ and $\hat{\bm{\mathrm{S}}}$ is written as $\mathcal{Z} (\hat{\bm{\mathrm{S}}}, \bm{\mathrm{X}})$ to distinguish between a raw OSQA score $\mathcal{Z} (\hat{\bm{\mathrm{S}}}, \bm{\mathrm{X}})$ and normalized OSQA score $\mathcal{B}(\hat{\bm{\mathrm{S}}},\bm{\mathrm{X}})$.

From (\ref{eq:total_obj}) and (\ref{eq:mean_obj}), 
the total gradient $\nabla _{\Theta} \mathcal{J}(\Theta)$ is a weighted sum of the $i$-th gradient of the log-likelihood function, and $\mathcal{B}(\hat{\bm{\mathrm{S}}},\bm{\mathrm{X}})$ is used as its weight.
Since typical OSQA scores vary not only by the performance of source enhancement but also by the SNRs of each input signal $\bm{\mathrm{X}}^{(1,...,I)}$, 
$\nabla _{\Theta} \mathcal{J}(\Theta)$ also varies by the OSQA scores and SNRs of $\bm{\mathrm{X}}^{(1,...,I)}$.
To reduce the variance in the estimate of the gradient, it would be better to remove such external factors according to the input conditions of each input signal, {\it e.g.}, input SNRs.
As a possible solution, the external factors involved in the OSQA score would be estimated by calculating the expectation of the OSQA score of the input signal.
Thus, subtracting the conditional expectation of $\mathcal{Z} (\hat{\bm{\mathrm{S}}}, \bm{\mathrm{X}})$ given by each input signal $\mathbb{E}_{\hat{\bm{\mathrm{S}}} | \bm{\mathrm{X}}} [ \mathcal{Z} ( \hat{\bm{\mathrm{S}}}, \bm{\mathrm{X}} ) ]$ from $\mathcal{Z} (\hat{\bm{\mathrm{S}}}, \bm{\mathrm{X}})$ might be effective in reducing the variance as 
\begin{align}
\mathcal{B} \left( \hat{\bm{\mathrm{S}}}, \bm{\mathrm{X}} \right) 
= \mathcal{Z}( \hat{\bm{\mathrm{S}}} , \bm{\mathrm{X}}) - \mathbb{E}_{\hat{\bm{\mathrm{S}}} | \bm{\mathrm{X}}} \left[ \mathcal{Z} ( \hat{\bm{\mathrm{S}}}, \bm{\mathrm{X}} ) \right].
\label{eq:advantage_reward}
\end{align}
This implementation is known as ``baseline-subtraction'' \cite{Policy_Grad,advantage_AC}.
Here, $\mathbb{E}_{\hat{\bm{\mathrm{S}}} | \bm{\mathrm{X}}} [ \mathcal{Z} ( \hat{\bm{\mathrm{S}}}, \bm{\mathrm{X}} ) ]$ cannot be analytically calculated, 
so we replace the expectation with the average of OSQA scores. 
Then the scoring function is designed as 
\begin{align}
\mathcal{B} \left( \hat{\bm{\mathrm{S}}}^{(i, k)}, \bm{\mathrm{X}}^{(i)} \right)
= 
\mathcal{Z}( \hat{\bm{\mathrm{S}}}^{(i,k)} , \bm{\mathrm{X}}^{(i)}) - \frac{1}{K} \sum_{j=1}^{K} \mathcal{Z} ( \hat{\bm{\mathrm{S}}}^{(i,j)}, \bm{\mathrm{X}}^{(i)} ).
\label{eq:calc_perceptual_score}
\end{align}

\subsection{Sampling-algorithm to simulate T-F-mask-processed output signal}
\label{sec:prop:C}
\begin{figure}[ttt]
\centering
\includegraphics[width=55mm]{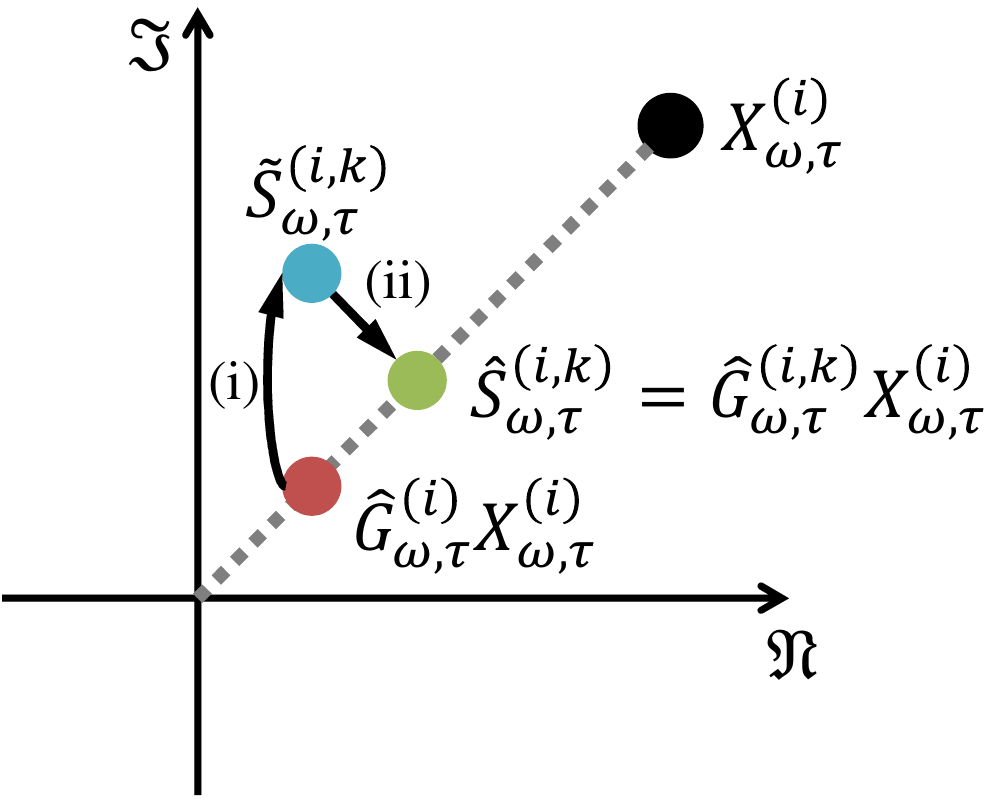}
\caption{T-F mask sampling procedure of proposed method on complex plane. 
The black, red, blue, and green points represent 
$X_{\omega, \tau}^{(i)}$,
$\hat{G}_{\omega, \tau}^{(i)} X_{\omega, \tau}^{(i)}$,
$\tilde{S}_{\omega, \tau}^{(i,k)}$,
and $\hat{G}_{\omega, \tau}^{(i,k)} X_{\omega, \tau}^{(i)}$, respectively.
First, the parameters of $p(\hat{S}_{\omega,\tau} | X_{\omega,\tau}^{(i)}, \Theta)$, {\it i.e.}, the T-F mask $\hat{G}_{\omega, \tau}^{(i)}$ and the variance are estimated using a DNN. 
Then, $\tilde{S}_{\omega, \tau}^{(i,k)}$ is sampled from $p(\hat{S}_{\omega,\tau} | X_{\omega,\tau}^{(i)}, \Theta)$ by using a typical sampling algorithm; which is shown as arrow-(i).
Finally, the simulated T-F mask $\hat{G}_{\omega, \tau}^{(i,k)}$ is calculated to minimize the MSE between $\tilde{S}_{\omega, \tau}^{(i,k)}$ and the simulated output signal $\hat{G}_{\omega, \tau}^{(i,k)} X_{\omega, \tau}^{(i)}$ by (\ref{eq:TF_gen}); which is shown as arrow-(ii).
}
\label{fig:TFmaskSampling}
\end{figure}

The sampling operator used in (\ref{eq:monte}) is an intuitive method that uses a typical pseudo random number generator such as the Mersenne-Twister \cite{Mersenne}. 
However, this sampling operator would in fact be difficult to use because typical sampling algorithms simulate output signals that do not satisfy the constraint of real-valued T-F-mask processing defined by (\ref{eq:TF-masking}). 
To avoid this problem, we calculate the T-F mask $\hat{G}_{\omega, \tau}^{(i,k)}$ and output signal $\hat{S}_{\omega,\tau}^{(i,k)}$ from the simulated output signal by using a typical sampling algorithm $\tilde{S}_{\omega, \tau}^{(i,k)}$, so that $\hat{G}_{\omega, \tau}^{(i,k)}$ and $\hat{S}_{\omega,\tau}^{(i,k)}$ satisfy the constraint of T-F-mask processing and minimize the squared error between $\hat{S}_{\omega,\tau}^{(i,k)}$ and $\tilde{S}_{\omega, \tau}^{(i,k)}$.

Figure \ref{fig:TFmaskSampling} illustrates the overview of the problem and the proposed solution on the complex plane. 
In this study, we use the real-value T-F mask within the range of $0 \leq G_{\omega, \tau} \leq 1$.
Thus, the output signal is constrained to exist on the dotted line in Fig. \ref{fig:TFmaskSampling}, \textit{i.e.}, T-F mask processing affects only the norm of $\hat{S}_{\omega,\tau}^{(i,k)}$. 
However, since $p(\hat{\bm{S}} | \bm{X}, \Theta )$ is modeled by a continuous PDF such as the complex Gaussian distribution in (\ref{eq:ML_obj}), a typical sampling algorithm possibly generates output signals that do not satisfy the T-F-mask constraint, \textit{i.e.}, the phase spectrum of $\tilde{S}_{\omega, \tau}^{(i,k)}$ does not coincide with that of $X_{\omega, \tau}^{(i)}$. 
To solve this problem, we formulate the PSA-based T-F-mask re-calculation. 
First, a temporary output signal $\tilde{S}_{\omega, \tau}^{(i,k)}$ is sampled using a sampling algorithm (Fig. \ref{fig:TFmaskSampling} arrow-(i)). 
Then, the T-F mask $\hat{G}_{\omega, \tau}^{(i,k)}$ that minimizes the squared error between $\tilde{S}_{\omega, \tau}^{(i,k)}$ and $\hat{G}_{\omega, \tau}^{(i,k)} X_{\omega, \tau}^{(i)}$ is calculated using the PSA equation as
\begin{align}
\hat{G}_{\omega, \tau}^{(i,k)} = \min\left( 1, \max \left( 0, \frac{|\tilde{S}_{\omega, \tau}^{(i,k)}|}{|X_{\omega, \tau}^{(i)}|} \cos \left( \theta_{\omega, \tau}^{(\tilde{S}^{(i,k)})} - \theta_{\omega, \tau}^{(X^{(i)})} \right) \right) \right), \label{eq:TF_gen}
\end{align}
where $\theta_{\omega, \tau}^{(\tilde{S}^{(i,k)})}$ and $\theta_{\omega, \tau}^{(X^{(i)})}$ are the phase spectra of $\tilde{S}_{\omega, \tau}^{(i,k)}$ and $X_{\omega, \tau}^{(i)}$, respectively. 
Then, the output signal is calculated by
\begin{equation}
\hat{S}_{\omega,\tau}^{(i,k)} = \hat{G}_{\omega,\tau}^{(i,k)} X_{\omega,\tau}^{(i)},
\label{eq:output_prop}
\end{equation}
as shown with arrow-(ii) in Fig. \ref{fig:TFmaskSampling}.

\subsection{Training procedure}
\label{sec:procedure}
\begin{figure*}[ttt]
\centering
\includegraphics[width=175mm]{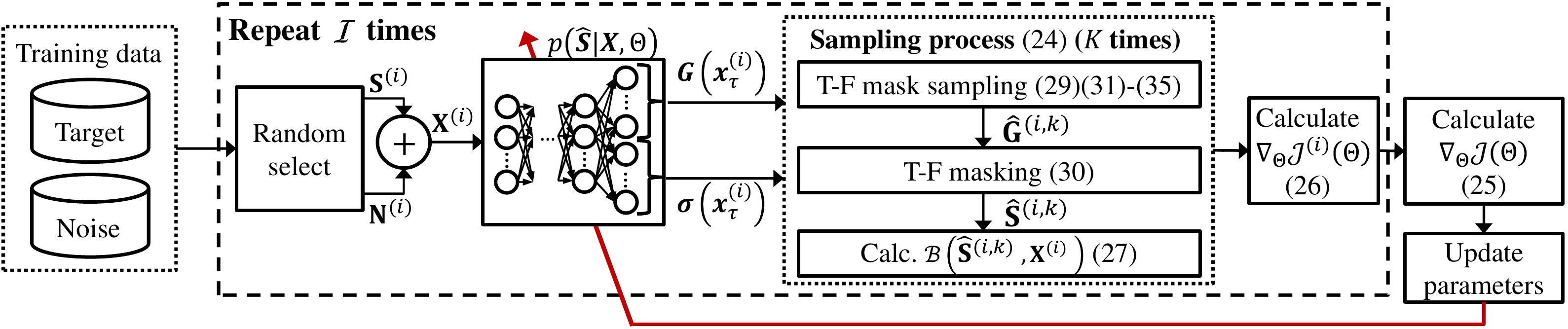}
\caption{Training procedure of proposed method}
\label{fig:procedure}
\end{figure*}

We describe the overall training procedure of the proposed method, as shown in Fig. \ref{fig:procedure}. 
Hereafter, to simplify the sampling algorithm, we use the complex Gaussian distribution as $p(\hat{\bm{S}} | \bm{X}, \Theta )$ described in (\ref{eq:ML_obj})--(\ref{eq:z_return}).

First, the $i$-th observation utterance $\bm{\mathrm{X}}^{(i)}$ is simulated by (\ref{eq:obs}) using a randomly selected target-source file and a noise source with equal frame size from the training dataset. Next, the T-F mask $\bm{G}(\bm{x}_{\tau}^{(i)})$ and variance $\bm{\sigma}(\bm{x}_{\tau}^{(i)})$ are estimated by (\ref{eq:ML_obj})--(\ref{eq:z_return}). Then, to simulate the $k$-th output signal $\hat{\bm{\mathrm{S}}}^{(i,k)}$, the temporary output signal $\tilde{S}_{\omega, \tau}^{(i,k)'}$ is sampled from the complex Gaussian distribution using a pseudo random number generator, such as the Mersenne-Twister \cite{Mersenne}, as 
\begin{align}
\begin{bmatrix}
\Re\left( \tilde{S}_{\omega, \tau}^{(i,k)} \right)\\
\Im\left( \tilde{S}_{\omega, \tau}^{(i,k)} \right)
\end{bmatrix}
&\sim
\mathcal{N}_{\mathbb{C}} \left( \hat{G}_{\omega, \tau}^{(i)}
\begin{bmatrix}
\Re\left( X_{\omega, \tau}^{(i)} \right)\\
\Im\left( X_{\omega, \tau}^{(i)} \right)
\end{bmatrix}
, \sigma^2_{\omega, \tau} \bm{I}
 \right), \label{eq:complex_gauss}
\end{align}
where $\bm{I}$ is the $2 \times 2$ identity matrix, and $\Re$ and $\Im$ denote the real and imaginary parts of the complex number, respectively. After that, T-F mask $\hat{G}_{\omega,\tau}^{(i,k)}$ is calculated using (\ref{eq:TF_gen}). To accelerate the algorithm convergence, we additionally use the $\epsilon$-greedy algorithm to calculate $\hat{G}_{\omega,\tau}^{(i,k)}$. With probability $1-\epsilon$ applied to each time-frequency bin, the maximum a posteriori (MAP) T-F mask $\hat{G}_{\omega, \tau}^{(i)}$ estimated using DNNs is used instead of $\hat{G}_{\omega, \tau}^{(i,k)}$ as
\begin{equation}
\hat{G}_{\omega, \tau}^{(i,k)} \gets 
\begin{cases}
\hat{G}_{\omega, \tau}^{(i,k)}	& ( \mbox{with prob. } \epsilon )\\
\hat{G}_{\omega, \tau}^{(i)}	& ( \mbox{otherwise} )
\end{cases}.
\end{equation}
In addition, a large gradient value $\nabla _{\Theta} \mathcal{J}(\Theta)$ leads to unstable training. One reason for the large gradient is that the log-likelihood $\nabla _{\Theta} \ln p(\hat{\bm{S}}_{\tau}^{(i, k)} | \bm{X}_{\tau}^{(i)}, \Theta )$ in (\ref{eq:mean_obj}) becomes large. To reduce the gradient of the log-likelihood, the difference between the mean T-F mask $\hat{G}_{\omega, \tau}^{(i)}$ and simulated T-F mask $\hat{G}_{\omega, \tau}^{(i,k)}$ is truncated to confine it within the range of $[-\lambda, \lambda]$ as
\begin{align}
\Delta \hat{G}_{\omega, \tau}^{(i,k)} &\gets \hat{G}_{\omega, \tau}^{(i,k)} - \hat{G}_{\omega, \tau}^{(i)}\\
\Delta \hat{G}_{\omega, \tau}^{(i,k)} &\gets 
\begin{cases}
\lambda									& ( \Delta \hat{G}_{\omega, \tau}^{(i,k)} > \lambda )\\
\Delta \hat{G}_{\omega, \tau}^{(i,k)}	& ( -\lambda \leq \Delta \hat{G}_{\omega, \tau}^{(i,k)} \leq \lambda )\\
-\lambda								& ( \Delta \hat{G}_{\omega, \tau}^{(i,k)} < -\lambda )
\end{cases},\\
\hat{G}_{\omega, \tau}^{(i,k)} &\gets \hat{G}_{\omega, \tau}^{(i)} + \Delta \hat{G}_{\omega, \tau}^{(i,k)}.
\end{align}
Then, the output signal $\hat{\bm{\mathrm{S}}}^{(i, k)}$ is calculated by T-F-mask processing (\ref{eq:output_prop}), and the OSQA scores $\mathcal{Z} ( \hat{\bm{\mathrm{S}}}^{(i, k)}, \bm{\mathrm{X}}^{(i)} )$ and $\mathcal{B} ( \hat{\bm{\mathrm{S}}}^{(i, k)}, \bm{\mathrm{X}}^{(i)} )$ are calculated by (\ref{eq:calc_perceptual_score}). After applying these procedures for $\mathcal{I}$ utterances, $\Theta$ is updated using the back-propagation algorithm using the gradient calculated by (\ref{eq:total_obj}).

\section{EXPERIMENTS}
\label{sec:eval}

We conducted objective experiments to evaluate the performance of the proposed method. 
The experimental conditions are described in Sec. \ref{sec:exp_cond}. 
To investigate whether a DNN source-enhancement function can be trained to increase OSQA scores, we first investigated the relationship between the number of updates and OSQA scores (Sec. \ref{sec:obj_eval_invst}). 
Second, the source enhancement performance of the proposed method was compared with those of conventional methods by using several objective measurements (Sec. \ref{sec:obj_eval}). 
Finally, subjective evaluations for sound quality and ineligibility were conducted (Sec. \ref{sec:eval_subj}). 
For comparison methods, we used four DNN source-enhancement methods; 
two T-F-mask mapping functions trained using an MMSE-based objective function \cite{Weninger_2015} and the ML-based objective function described in Sec. \ref{sec:ml_dnn}, 
and two T-F-mask selection functions trained for increasing the PESQ and STOI \cite{Koizumi_ICASSP_2017}.

\subsection{Experimental conditions}
\label{sec:exp_cond}
\begin{table}[tt]
\centering
\caption{Experimental conditions}
\small
\begin{tabular}{l|l}\hline \hline
	\multicolumn{2}{c}{Parameters for signal processing}\\ \hline 
 Sampling rate							& 16.0 kHz \\
 FFT length			 					& 512 pts	\\
 FFT shift length	 					& 256 pts	\\ 
 \# of mel-filterbanks 					& 64 		\\ 
 Smoothing parameter	$\beta$ 			& 0.3 \\ 
 Lower threshold	$G^{\mbox{\scriptsize min}}$ 	& 0.158 $(=-16 \mbox{ dB})$ \\ 
 Training SNR (dB)	 	 				& -6, 0, 6, 12	\\ \hline
	\multicolumn{2}{c}{DNN architecture}\\ \hline 
 \# of hidden layers for DNNs 			& $3$ \\ 
 \# of hidden units for DNNs 			& $1024$ \\ 
 Activation function	(T-F mask, $\phi_{g}$)			& sigmoid \\ 
 Activation function	(variance, $\phi_{\sigma}$)		& exponential \\ 
 Activation function	(hidden, $\phi_{h}$)			& ReLU \\ 
 Context window size	$Q$					& 5 \\ 
 Variance regularization parameter $C_{\sigma}$ & $10^{-4}$ \\ \hline
	\multicolumn{2}{c}{Parameters for MMSE and ML-based DNN training}\\ \hline 
 Initial step-size						& $10^{-4}$	\\ 
 Step-size threshold for early-stopping	& $10^{-7}$	\\ 
 Dropout probability (input layer) 		& 0.2	\\ 
 Dropout probability (hidden layer) 		& 0.5	\\ 
	$L_2$ normalization parameter			& $10^{-4}$ \\ \hline 
	\multicolumn{2}{c}{Parameters for T-F mask selection}\\ \hline 
 \# of T-F mask templates						& 128  \\ 
 $\epsilon$-greedy parameter $\epsilon$			& 0.01 \\ \hline 
	\multicolumn{2}{c}{Parameters for proposed DNN training}\\ \hline 
 Step-size 								& $10^{-6}$	\\ 
 \# of utterance	$\mathcal{I}$ 					& 10 \\ 
 \# of T-F mask sampling	$K$						& 20 \\ 
 Clipping parameter	$\lambda$ 					& 0.05 \\ 
 $\epsilon$-greedy parameter $\epsilon$			& 0.05 \\ \hline \hline
\end{tabular}
\normalsize
\label{tbl:param_tbl}
\end{table}

\begin{figure*}[tttt]
\centering
\includegraphics[width=170mm]{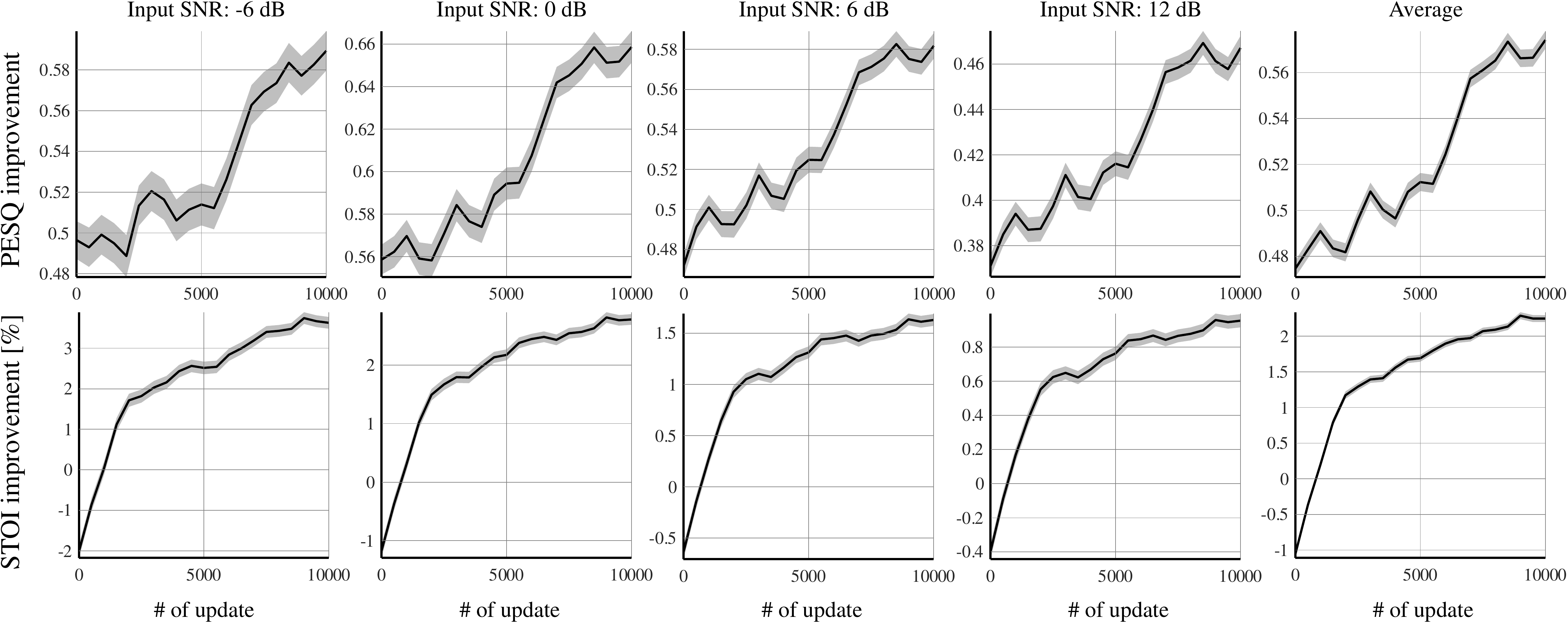}
\caption{
OSQA score improvement depending on number of updates. 
X-axis shows number of updates, and y-axis shows average difference between OSQA score of proposed method and that of observed signal. 
Solid lines and gray area are average and standard-error, respectively. }
\label{fig:obj_result}
\vspace{10pt}
\centering
\includegraphics[width=175mm]{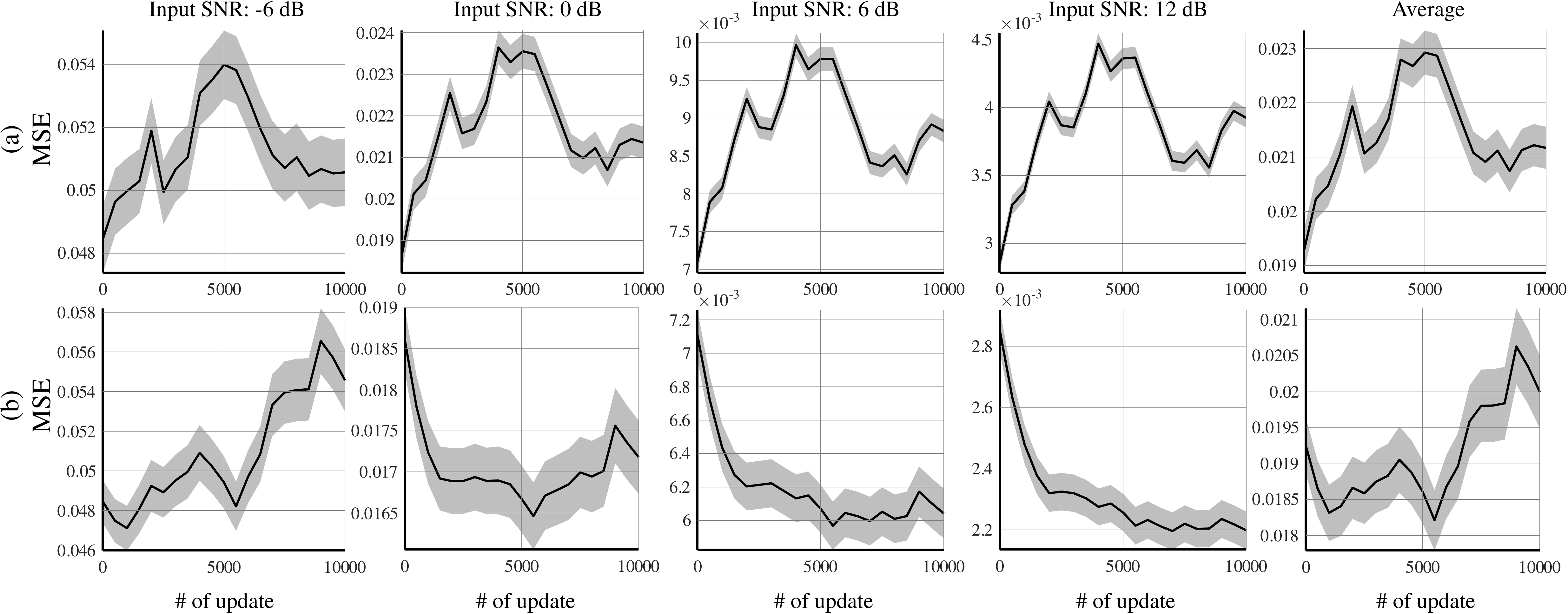}
\caption{
Mean squared error (MSE) depending on number of updates.
OSQA scores used for training of proposed method were (a) PESQ and (b) STOI. 
X-axis shows number of updates, and y-axis shows MSE. 
Solid lines and gray area are average and standard-error, respectively. }
\label{fig:obj_result_mse}
\end{figure*}

\vspace{5pt}
\subsubsection{Dataset}

The ATR Japanese speech database \cite{ATR_database} was used as the training dataset of the target source. 
The dataset consists of 6640 utterances spoken by 11 males and 11 females. 
The utterances were randomly separated into 5976 for the development set and 664 for the validation set. 
As the training dataset of noise, a noise dataset of CHiME-3 was used that consisted of four types of background noise files including noise in {\it cafes}, {\it street junctions}, {\it public transport}, and {\it pedestrian areas} \cite{CHiME}. 
The noisy-mixture dataset was generated by mixing clean speech utterances with various noisy and SNR conditions using the following procedure;
i) the noise is randomly selected from noise dataset, 
ii) the amplitude of noise is adjusted to be the desired SNR-level,
and iii) the speech and noise source is added in the time-domain.
As the test dataset, a Japanese speech database consisting of 300 utterances spoken by 3 males and 3 females was used for target-source dataset, and an ambient noise database recorded at 
{\it airports} (Airp.), {\it amusement parks} (Amuse.), {\it offices} (Office), and {\it party rooms} (Party) was used as the noisy dataset. 
All samples were recorded at the sampling rate of 16 kHz. 
The SNR levels of the training/test dataset were -6, 0, 6, and 12 dB.

\vspace{5pt}
\subsubsection{DNN architecture and setup}

For the proposed and all conventional methods, a fully connected DNN was used that has 3 hidden layers and 1024 hidden units. 
All input vectors were mean-and-variance normalized using the training data statistics. 
The activation functions for the T-F mask $\phi_{g}$, variance $\phi_{\sigma}$, and hidden units $\phi_{h}$ were the sigmoid function, exponential function, and rectified linear unit (ReLU), respectively. 
The context window size was $Q = 5$, and the variance regularization parameter in (\ref{eq:DNN-sigma}) was $C_{\sigma} = 10^{-4}$\footnote{
In preliminary experiments using candidate values $C_{\sigma} \in \{ 10^{-2}, 10^{-3}, 10^{-4} \}$, there were no distinct differences in training stability and results. 
Thus, to eliminate the effect of regularization, we used the minimum parameter of the candidate values.
}
. 
The Adam method \cite{adam} was used as a gradient method. 
To avoid over-fitting, input vectors and DNN outputs, {\it i.e.}, the T-F masks and error variances, were compressed using a $\mathsf{B}=64$ Mel-transformation matrix, and the estimated T-F masks and error variances were transformed into a linear frequency domain using the Mel-transform's pseudo-inverse \cite{Weninger_2014}.

A PSA objective function \cite{Weninger_2015,Erdogan_2015} was used as the MMSE-based objective function. 
Since the PSA objective function does not use the variance parameter $\bm{\sigma}(\bm{x}_{\tau})$, DNNs estimate only T-F masks $\bm{G}(\bm{x}_{\tau})$. 
For the ML-based objective function, we used (\ref{eq:ml_obj_sum}) with the complex Gaussian distribution described in Sec. \ref{sec:ml_dnn}. 
To train both methods, the dropout algorithm was used and initialized by layer-by-layer pre-training \cite{pre-train}. 
An early-stopping algorithm \cite{Koizumi_2017} was used for fine-tuning with the initial step-size $10^{-4}$ and the step-size threshold $10^{-7}$, and L2 normalization with the parameter $10^{-4}$ was used as a regularization algorithm.

For the T-F-mask selection-based method \cite{Koizumi_ICASSP_2017}, to improve the flexibility of T-F-mask selection, we used 128 T-F-mask templates.
The DNN architecture, except for the output layer, is the same as MMSE- and ML-based methods.

For the proposed method, DNN parameters were initialized by ML-based training, and their step-size was $10^{-6}$. 
To calculate $\nabla_{\Theta} \mathcal{J}(\Theta)$, the iteration parameters $\mathcal{I} = 10$ and $K = 20$ were used. 
The $\epsilon$-greedy parameter $\epsilon$ was 0.05, and the clipping parameter $\lambda$ was determined as $0.05$ according to preliminary informal experiments\footnote{
We tested some possible combinations of these parameters by grid-search.
Then, we found that the listed parameters achieved a stable training and realistic computational time (2 days using an Intel Xeon Processor E5-2630 v3 CPU and a Tesla M-40 GPU).
}.
As the OSQA scores, we used the PSEQ, which is a speech quality measure, and the STOI, which is a speech intelligibility measure. 
To avoid adjusting the step-size of the gradient method for each OSQA, we normalized OSQA scores to uniform the range of the each OSQA score.
In this experiments, each OSQA score was normalized so that its maximum and minimum values were 100 and 0 as
\begin{align}
\nonumber
\mathcal{Z}^{\mbox{\scriptsize PESQ}}(\hat{\bm{\mathrm{S}}}, \bm{\mathrm{X}}) &= 20.0 \times  \left( \mbox{PESQ}(\hat{\bm{\mathrm{S}}}, \bm{\mathrm{X}}) + 0.5 \right) , \\
\nonumber
\mathcal{Z}^{\mbox{\scriptsize STOI}}(\hat{\bm{\mathrm{S}}}, \bm{\mathrm{X}}) &= 100.0 \times \mbox{STOI}(\hat{\bm{\mathrm{S}}}, \bm{\mathrm{X}}) .
\end{align}
The training algorithm was stopped after 10,000 times of executing the whole parameter update process shown in Fig. \ref{fig:procedure}.

\vspace{5pt}
\subsubsection{Other conditions}

It is known that T-F-mask processing causes artificial distortion, so-called musical noise \cite{Miyazaki_2012}. For all methods, to reduce musical noise, flooring \cite{Lightburn_2017,Cohen_2002} and smoothing \cite{Emmanuel_2010,Niwa_IWAENC} were applied to $\hat{G}_{\omega, \tau}$ before T-F-mask processing as 
\begin{align}
\hat{G}_{\omega, \tau} &\gets \max \left( G^{\mbox{\scriptsize min}}, \hat{G}_{\omega, \tau} \right) ,\\
\hat{G}_{\omega, \tau} &\gets \beta \hat{G}_{\omega, \tau} + (1-\beta) \hat{G}_{\omega, \tau -1} ,
\end{align}
where we used the lower threshold of the T-F mask $G^{\mbox{\scriptsize min}} = 0.158$ and smoothing parameter $\beta = 0.3$. 
The frame size of the short-time Fourier transform (STFT) was 512, and the frame was shifted by 256 samples. All the above-mentioned conditions are summarized in Table \ref{tbl:param_tbl}.

\subsection{Investigation of relationship between number of updates and OSQA score}
\label{sec:obj_eval_invst}

\begin{table}[ttt]
\centering
\caption{Correlation coefficients between MSE and OSQA score improvements}
\begin{tabular}{l|ccccc} \hline \hline
		 &	-6 dB		&	0 dB		&	6 dB		&	12 dB		&	Average		\\ \hline
	PESQ &	$-0.120$	&	$-0.081$	&	$0.020$		&	$0.089$		&	$-0.020$	 \\ 
	STOI &	$0.756$		&	$-0.672$	&	$-0.951$	&	$-0.980$	&	$0.482$	 \\ \hline \hline
\end{tabular}
\label{tbl:corrCoeff}
\end{table}

\begin{table*}[ttt]
  \centering
  \caption{Evaluation results on three objective measurements.
Asterisks indicate scores significantly higher than that of {\tt MMSE} and {\tt ML} in paired one-sided t-test. 
Gray cells indicate the highest score in same noise and input SNR condition.
}
\footnotesize
  Input SNR: -6 dB\\
\scriptsize
  \begin{tabular}{l|cccc|c|cccc|c|cccc|c} \hline \hline
							&	\multicolumn{5}{c}{SDR [dB]} &	\multicolumn{5}{|c|}{PESQ} &	\multicolumn{5}{c}{STOI [\%]} \\ \hline
	Method					& 	Airp.	& Amuse. &	Office & Party & Ave.
							& 	Airp.	& Amuse. &	Office & Party & Ave. 
							&	Airp.	& Amuse. &	Office & Party & Ave.\\ \hline
 	{\tt OBS}
							&	$-4.28$        & $-6.98$        & $-5.64$        & $-1.50$        & $-4.6$ 
							& 	$1.24$        & $1.38$        & $1.33$        & $1.14$        & $1.27$     
							&	$72.1$        & $76.7$        & $73.8$        & $69.1$        & $72.9$    \\
 	{\tt MMSE}
							&	$3.22$        & $5.87$        & $4.66$        & $3.77$        & $4.38$             
							& 	$1.66$        & $1.89$        & $1.80$        & $1.48$        & $1.71$ 
							&	$68.9$        & $73.6$        & $71.0$        & $66.7$        & $70.1$        \\
	{\tt ML}
							&	\cellcolor[gray]{0.8} $\bf{3.31}$   & $6.12$        & \cellcolor[gray]{0.8} $\bf{4.87}$   & $3.63$        & \cellcolor[gray]{0.8} $\bf{4.48}$        
							& 	$1.68$        & $1.95$        & $1.80$        & $1.54$        & $1.74$   
							&	$69.2$        & $74.3$        & $72.0$        & $64.9$        & $70.1$    \\ \hline
	{\tt C-PESQ} 
							&	$-0.28$        & $1.38$        & $-0.03$        & $1.67$        & $0.69$          
							& 	$1.55$        & $1.77$        & $1.64$        & $1.44$        & $1.60$  
							&	$*72.2$        & $*76.4$        & $*73.4$        & $*70.4$        & $*73.2$    \\
	{\tt C-STOI} 
							&	$0.21$        & $2.02$        & $0.68$        & $2.17$        & $1.27$            
							& 	$1.48$        & $1.64$        & $1.56$        & $1.34$        & $1.50$   
							&	\cellcolor[gray]{0.8} $\bf{*75.0}$   & $*79.8$        & \cellcolor[gray]{0.8} $\bf{*76.6}$   & $*71.1$        & $*75.6$        \\  \hline
	{\tt P-PESQ}
							&	$3.13$        & $*6.34$        & $4.72$        & $3.50$        & $4.42$ 
							& 	\cellcolor[gray]{0.8} $\bf{*1.78}$   & $*2.07$        & \cellcolor[gray]{0.8} $\bf{*1.91}$   & $*1.57$        & \cellcolor[gray]{0.8} $\bf{*1.83}$   
							&	$*71.0$        & $*76.0$        & $*72.4$        & $*67.9$        & $*71.8$         \\
	{\tt P-STOI}
							&	$2.18$        & \cellcolor[gray]{0.8} $\bf{*6.60}$   & $3.90$        & \cellcolor[gray]{0.8} $\bf{*4.15}$   & $4.21$           
							& 	$1.63$        & $1.93$        & $1.73$        & \cellcolor[gray]{0.8} $\bf{*1.59}$   & $1.72$    
							&	$*74.9$        & \cellcolor[gray]{0.8} $\bf{*80.1}$   & \cellcolor[gray]{0.8} $\bf{*76.6}$   & \cellcolor[gray]{0.8} $\bf{*71.3}$   & \cellcolor[gray]{0.8} $\bf{*75.7}$    \\ 
	{\tt P-MIX} 
							&	$2.93$        & $*6.20$        & $4.39$        & $3.49$        & $4.25$         
							& 	$*1.77$        & \cellcolor[gray]{0.8} $\bf{*2.08}$   & $*1.89$        & \cellcolor[gray]{0.8} $\bf{*1.59}$         & \cellcolor[gray]{0.8} $\bf{*1.83}$    
							&	$*72.1$        & $*77.4$        & $*73.8$        & $*68.2$        & $*72.9$          \\ 
  \hline \hline
  \end{tabular}\\
  \vspace{10pt}
\footnotesize
  Input SNR: 0 dB\\
\scriptsize
  \begin{tabular}{l|cccc|c|cccc|c|cccc|c} \hline \hline
							&	\multicolumn{5}{c}{SDR [dB]} &	\multicolumn{5}{|c|}{PESQ} &	\multicolumn{5}{c}{STOI [\%]} \\ \hline
	Method					& 	Airp.	& Amuse. &	Office & Party & Ave.
							& 	Airp.	& Amuse. &	Office & Party & Ave. 
							&	Airp.	& Amuse. &	Office & Party & Ave.\\ \hline
 	{\tt OBS}
							&	$1.67$        & $-1.19$        & $0.36$        & $4.46$        & $1.32$     
							& 	$1.71$        & $1.88$        & $1.81$        & $1.54$        & $1.73$  
							&	$84.5$        & $87.8$        & $85.2$        & $82.9$        & $85.1$   \\
 	{\tt MMSE}
							&	$8.03$        & $10.0$        & $9.55$        & $8.44$        & $9.00$            
							& 	$2.17$        & $2.36$        & $2.27$        & $2.09$        & $2.22$    
							&	$80.7$        & $84.7$        & $83.1$        & $80.1$        & $82.1$       \\
	{\tt ML}
							&	\cellcolor[gray]{0.8} $\bf{8.62}$   & $10.4$        & \cellcolor[gray]{0.8} $\bf{9.97}$   & $8.66$        & $9.40$             
							& 	$2.20$        & $2.42$        & $2.30$        & $2.14$        & $2.27$     
							&	$82.5$        & $86.4$        & $84.6$        & $79.6$        & $83.3$    \\ \hline
	{\tt C-PESQ} 
							&	$6.36$        & $7.08$        & $6.49$        & $7.89$        & $6.95$             
							& 	$2.11$        & $2.33$        & $2.23$        & $2.00$        & $2.16$    
							&	$*83.7$        & $86.2$        & $84.0$        & $*82.7$        & $*84.2$        \\
	{\tt C-STOI} 
							&	$7.30$        & $8.07$        & $7.18$        & $8.70$        & $7.81$               
							& 	$2.03$        & $2.18$        & $2.10$        & $1.89$        & $2.05$  
							&	\cellcolor[gray]{0.8} $\bf{*86.8}$   & $*89.9$        & $*87.4$        & $*84.7$        & $*87.2$  \\  \hline
	{\tt P-PESQ}
							&	$8.40$        & $10.3$        & $9.77$        & $8.28$        & $9.19$    
							& 	$*2.30$        & $*2.55$        & \cellcolor[gray]{0.8} $\bf{*2.41}$   & $*2.20$        & $*2.37$   
							&	$*82.7$        & $86.4$        & $84.1$        & $*80.3$        & $*83.4$   \\
	{\tt P-STOI}
							&	$8.45$        & \cellcolor[gray]{0.8} $\bf{*11.2}$   & $9.52$        & \cellcolor[gray]{0.8} $\bf{*9.74}$   & \cellcolor[gray]{0.8} $\bf{*9.74}$          
							& 	$2.12$        & $2.36$        & $2.21$        & $2.11$        & $2.20$   
							&	$*86.7$        & \cellcolor[gray]{0.8} $\bf{*90.0}$   & \cellcolor[gray]{0.8} $\bf{*87.5}$   & \cellcolor[gray]{0.8} $\bf{*85.0}$   & \cellcolor[gray]{0.8} $\bf{*87.3}$     \\ 
	{\tt P-MIX} 
							&	$8.09$        & $9.85$        & $9.12$        & $8.11$        & $8.79$          
							& 	\cellcolor[gray]{0.8} $\bf{*2.31}$   & \cellcolor[gray]{0.8} $\bf{*2.57}$   & $*2.41$        & \cellcolor[gray]{0.8} $\bf{*2.23}$   & \cellcolor[gray]{0.8} $\bf{*2.38}$   
							&	$*84.2$        & $*87.8$        & $*85.5$        & $*81.6$        & $*84.7$        \\ 
  \hline \hline
  \end{tabular}\\
  \vspace{10pt}
\footnotesize
  Input SNR: 6 dB\\
\scriptsize
  \begin{tabular}{l|cccc|c|cccc|c|cccc|c} \hline \hline
							&	\multicolumn{5}{c}{SDR [dB]} &	\multicolumn{5}{|c|}{PESQ} &	\multicolumn{5}{c}{STOI [\%]} \\ \hline
	Method					& 	Airp.	& Amuse. &	Office & Party & Ave.
							& 	Airp.	& Amuse. &	Office & Party & Ave. 
							&	Airp.	& Amuse. &	Office & Party & Ave.\\ \hline
 	{\tt OBS}
							&	$7.67$        & $4.96$        & $6.29$        & $10.5$        & $7.34$     
							& 	$2.18$        & $2.33$        & $2.28$        & $2.02$        & $2.20$   
							&	$92.2$        & $93.8$        & $92.7$        & $91.8$        & $92.6$     \\
 	{\tt MMSE}
							&	$12.1$        & $13.6$        & $13.4$        & $12.6$        & $12.9$             
							& 	$2.54$        & $2.68$        & $2.63$        & $2.49$        & $2.58$    
							&	$88.9$        & $91.2$        & $90.4$        & $88.6$        & $89.8$         \\
	{\tt ML}
							&	$13.1$        & $14.2$        & $14.1$        & $13.5$        & $13.7$          
							& 	$2.59$        & $2.77$        & $2.69$        & $2.54$        & $2.65$  
							&	$91.1$        & $93.0$        & $92.2$        & $89.8$        & $91.5$        \\ \hline
	{\tt C-PESQ} 
							&	$11.5$        & $11.9$        & $11.4$        & $12.6$        & $11.9$     
							& 	$2.54$        & $2.75$        & $2.69$        & $2.45$        & $2.61$ 
							&	$90.5$        & $91.8$        & $90.9$        & $89.9$        & $90.8$     \\
	{\tt C-STOI} 
							&	$13.2$        & $13.6$        & $13.1$        & $14.3$        & $13.5$         
							& 	$2.50$        & $2.62$        & $2.57$        & $2.38$        & $2.52$  
							&	\cellcolor[gray]{0.8} $\bf{*93.4}$   & $*94.8$        & $*93.9$        & \cellcolor[gray]{0.8} $\bf{*92.8}$   & \cellcolor[gray]{0.8} $\bf{*93.8}$  \\  \hline
	{\tt P-PESQ}
							&	$12.6$        & $13.8$        & $13.6$        & $12.6$        & $13.2$  
							& 	\cellcolor[gray]{0.8} $\bf{*2.70}$   & $*2.89$        & \cellcolor[gray]{0.8} $\bf{*2.80}$   & $*2.64$        & \cellcolor[gray]{0.8} $\bf{*2.76}$       
							&	$90.2$        & $92.1$        & $91.2$        & $89.1$        & $90.6$   \\
	{\tt P-STOI}
							&	\cellcolor[gray]{0.8} $\bf{*13.4}$   &\cellcolor[gray]{0.8}  $\bf{*15.3}$   & \cellcolor[gray]{0.8} $\bf{*14.3}$   & \cellcolor[gray]{0.8} $\bf{*14.8}$   & \cellcolor[gray]{0.8} $\bf{*14.4}$      
							& 	$2.49$        & $2.69$        & $2.60$        & $2.45$        & $2.56$        
							&	\cellcolor[gray]{0.8} $\bf{*93.4}$   & \cellcolor[gray]{0.8} $\bf{*94.9}$   & \cellcolor[gray]{0.8} $\bf{*94.0}$   & \cellcolor[gray]{0.8} $\bf{*92.8}$   & \cellcolor[gray]{0.8} $\bf{*93.8}$       \\ 
	{\tt P-MIX} 
							&	$11.5$        & $12.3$        & $12.1$        & $11.6$        & $11.9$    
							& 	$*2.69$        &\cellcolor[gray]{0.8} $\bf{*2.90}$   & $*2.79$        & \cellcolor[gray]{0.8} $\bf{*2.66}$   & \cellcolor[gray]{0.8} $\bf{*2.76}$  
							&	$*91.5$        & $*93.1$        & $*92.3$        & $*90.4$        & $*91.8$    \\ 
  \hline \hline
  \end{tabular}\\
  \vspace{10pt}
\footnotesize
  Input SNR: 12 dB\\
\scriptsize
  \begin{tabular}{l|cccc|c|cccc|c|cccc|c} \hline \hline
							&	\multicolumn{5}{c}{SDR [dB]} &	\multicolumn{5}{|c|}{PESQ} &	\multicolumn{5}{c}{STOI [\%]} \\ \hline
	Method					& 	Airp.	& Amuse. &	Office & Party & Ave.
							& 	Airp.	& Amuse. &	Office & Party & Ave. 
							&	Airp.	& Amuse. &	Office & Party & Ave.\\ \hline
 	{\tt OBS}
							&	$13.6$        & $11.0$        & $12.3$        & $16.4$        & $13.3$    
							& 	$2.61$        & $2.76$        & $2.72$        & $2.47$        & $2.64$    
							&	$96.1$        & $96.9$        & $96.4$        & $96.2$        & $96.4$     \\
 	{\tt MMSE}
							&	$15.9$        & $16.9$        & $16.8$        & $16.3$        & $16.5$             
							& 	$2.84$        & $2.95$        & $2.92$        & $2.77$        & $2.87$   
							&	$93.5$        & $94.7$        & $94.4$        & $93.2$        & $94.0$      \\
	{\tt ML}
							&	$17.5$        & $18.0$        & $18.0$        & $18.1$        & $17.9$           
							& 	$2.95$        & $3.09$        & $3.03$        & $2.88$        & $2.98$  
							&	$95.5$        & $96.3$        & $96.0$        & $94.9$        & $95.7$   \\ \hline
	{\tt C-PESQ} 
							&	$15.5$        & $15.8$        & $15.3$        & $16.3$        & $15.7$            
							& 	$2.95$        & $*3.14$        & $*3.08$        & $2.86$        & $*3.01$    
							&	$94.2$        & $94.9$        & $94.4$        & $94.0$        & $94.4$        \\
	{\tt C-STOI} 
							&	\cellcolor[gray]{0.8} $\bf{*18.2}$   & $*18.6$        & $*18.2$        & $*19.0$        & $*18.5$            
							& 	$2.94$        & $3.05$        & $3.01$        & $2.81$        & $2.95$     
							&	$*96.7$        & $*97.4$        & $*97.0$        & $*96.6$        & $*96.9$     \\  \hline
	{\tt P-PESQ}
							&	$16.5$        & $17.2$        & $17.1$        & $16.6$        & $16.8$   
							& 	\cellcolor[gray]{0.8} $\bf{*3.04}$   & \cellcolor[gray]{0.8} $\bf{*3.19}$   &\cellcolor[gray]{0.8} $\bf{*3.12}$   & \cellcolor[gray]{0.8} $\bf{*2.97}$        & \cellcolor[gray]{0.8} $\bf{*3.08}$  
							&	$94.4$        & $95.2$        & $94.9$        & $93.8$        & $94.6$   \\
	{\tt P-STOI}
							&	\cellcolor[gray]{0.8} $\bf{*18.2}$        & \cellcolor[gray]{0.8} $\bf{*19.5}$   & \cellcolor[gray]{0.8} $\bf{*18.8}$   & \cellcolor[gray]{0.8} $\bf{*19.7}$   & \cellcolor[gray]{0.8} $\bf{*19.1}$       
							& 	$2.85$        & $3.02$        & $2.96$        & $2.78$        & $2.90$   
							&	\cellcolor[gray]{0.8} $\bf{*96.8}$   & \cellcolor[gray]{0.8} $\bf{*97.5}$   &\cellcolor[gray]{0.8}  $\bf{*97.1}$   & \cellcolor[gray]{0.8} $\bf{*96.7}$   & \cellcolor[gray]{0.8} $\bf{*97.0}$   \\ 
	{\tt P-MIX} 
							&	$13.6$        & $13.9$        & $13.9$        & $13.8$        & $13.8$     
							& 	$*3.01$        & $*3.18$        & $*3.10$        & \cellcolor[gray]{0.8} $\bf{*2.97}$   & $*3.07$   
							&	$95.3$        & $96.0$        & $95.7$        & $94.7$        & $95.4$       \\ 
  \hline \hline
  \end{tabular}\\
  \label{tbl:obj_result_sdr}
\end{table*}

To investigate whether the DNN source-enhancement function can be trained to increase OSQA scores, we first investigated the relationship between the number of updates and improvement of the OSQA scores. 
We define ``OSQA score improvement'' as the difference in the score value from the baseline OSQA score. 
For the baseline, we use the OSQA score obtained from the observed signal.
Since the DNN parameters of the proposed method were initialized by ML-based training, each OSQA score was compared with the OSQA score that had zero updates. 
Thus, if DNN parameters were successfully trained with the proposed method, the OSQA score improvement would increase in accordance with the number of updates.

Figure \ref{fig:obj_result} shows the OSQA score improvements evaluated on the test dataset. 
Both OSQA score improvements increased as the number of updates increased for all SNR conditions. 
These results suggest that the proposed method is effective at increasing arbitrary OSQA scores, such as the PESQ and STOI.

We also investigated the relationship between the number of updates and MSE using the test dataset. 
Figure \ref{fig:obj_result_mse} shows MSE depending on the number of updates. 
Under most SNR conditions, MSE did not decrease despite OSQA scores increasing. 
Table \ref{tbl:corrCoeff} shows the correlation coefficients between OSQA score improvements and MSE values. 
There was little correlation between PESQ improvement and MSE, and the correlation between STOI improvement and MSE depended on the input SNR condition.
Thus, these results suggest that minimization of MSE does not necessarily maximize OSQA scores.

\subsection{Objective evaluation}
\label{sec:obj_eval}

\begin{table}[ttt]
\centering
\caption{Objective scores of example results shown in Fig. \ref{fig:result_example}.}
\begin{tabular}{l|cccc} \hline \hline
							&\multicolumn{3}{c}{Performance measurement}\\ 
	Method				&	SDR [dB]											&	PESQ 												&	STOI [\%] \\ \hline
	{\tt OBS}				&	$ 2.36$											&	$ 1.79$											&	$81.5$ \\ 
	{\tt MMSE}			&	$ 9.31$											&	$ 2.32$											&	$80.0$ \\ 
	{\tt ML}				&	\cellcolor[gray]{0.8} $\bm{11.3}$	&	$ 2.48$											&	$82.1$ \\ 
	{\tt P-PESQ}		&	$ 10.7$											&	\cellcolor[gray]{0.8} $\bm{2.55}$	&	$81.4$ \\ 
	{\tt P-STOI}		 	&	$ 11.2$											&	$ 2.40$											&	\cellcolor[gray]{0.8} $\bm{86.3}$ \\ 
	{\tt P-MIX}		 	&	$ 11.2$											&	\cellcolor[gray]{0.8} $\bm{2.55}$ &	$83.4$ \\ \hline \hline
\end{tabular}
\label{tbl:example}
\end{table}

\begin{figure*}[hhhh]
\centering
\includegraphics[width=170mm]{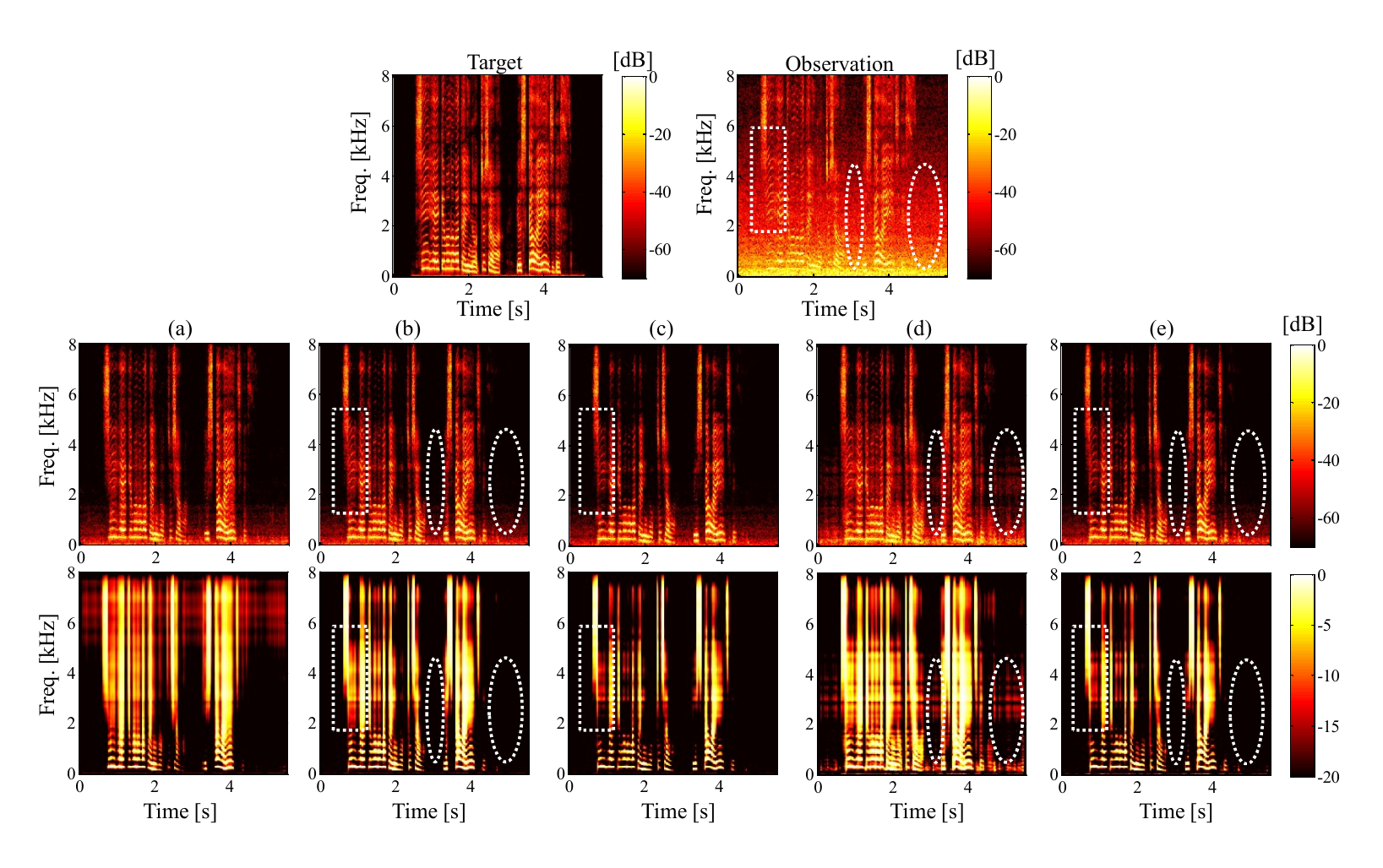}
\caption{Examples of estimated T-F mask and output signal. 
Top figures show spectrogram of target source $S_{\omega, \tau}$ (left) and observed signal $X_{\omega, \tau}$ (right), respectively. 
Middle figures show spectrogram of output signal $\hat{S}_{\omega, \tau}$ and bottom figures show estimated T-F mask $\hat{G}_{\omega, \tau}$, respectively. 
White dotted box and circle show larger or less noise reduction areas which modified by training of {\tt P-PESQ} and {\tt P-STOI}, respectively.
(a) {\tt MMSE}, (b) {\tt ML}, (c) {\tt P-PESQ}, (d) {\tt P-STOI}, and (e) {\tt P-MIX}. }
\label{fig:result_example}
\end{figure*}

The source-enhancement performance of the proposed method was compared with those of conventional methods using three objective measurements: the signal-to-distortion ratio (SDR), PESQ, and STOI. 
The SDR was defined as
\begin{equation}
\mbox{SDR [dB]} := 10 \log _{10} \frac{\sum_{\tau=1}^{T}\sum_{\omega=1}^{\Omega} |S_{\omega, \tau}|^2}{\sum_{\tau=1}^{T}\sum_{\omega=1}^{\Omega} |S_{\omega, \tau} - \hat{S}_{\omega, \tau}|^2},
\end{equation}
and calculated using the ``BSS-Eval toolbox \cite{Vincent2006}.''
These measurements were evaluated on the observed signal ({\tt OBS}), the MMSE- and ML-based DNN training ({\tt MMSE} and {\tt ML}), a T-F-mask selection method to increase the PESQ and STOI \cite{Koizumi_ICASSP_2017} ({\tt C-PESQ} and {\tt C-STOI}), and the proposed method to increase the PESQ and STOI ({\tt P-PESQ} and {\tt P-STOI}).
To investigate whether the proposed method enables training of a DNN to increase a metric that consists of multiple OSQA scores, we also trained a DNN to increase a mixed-OSQA score ({\tt P-MIX}).
As the first trial, we mixed the PESQ and the STOI.
The mixed-OSQA is defined as
\begin{equation}
\nonumber
\mathcal{Z}^{\mbox{\scriptsize MIX}}(\hat{\bm{\mathrm{S}}}, \bm{\mathrm{X}}) =
\gamma \mathcal{Z}^{\mbox{\scriptsize PESQ}}(\hat{\bm{\mathrm{S}}}, \bm{\mathrm{X}}) + 
\left( 1 - \gamma \right)
\mathcal{Z}^{\mbox{\scriptsize STOI}}(\hat{\bm{\mathrm{S}}}, \bm{\mathrm{X}}).
\end{equation}
In this trial, in order to confirm whether multiple OSQA scores increase simultaneously, the additive coefficient $\gamma = 0.5$ was determined in such a way that both OSQA scores had the same contribution to $\mathcal{Z}^{\mbox{\scriptsize MIX}}(\hat{\bm{\mathrm{S}}}, \bm{\mathrm{X}})$.

\begin{figure*}[tttt]
\centering
\includegraphics[width=155mm]{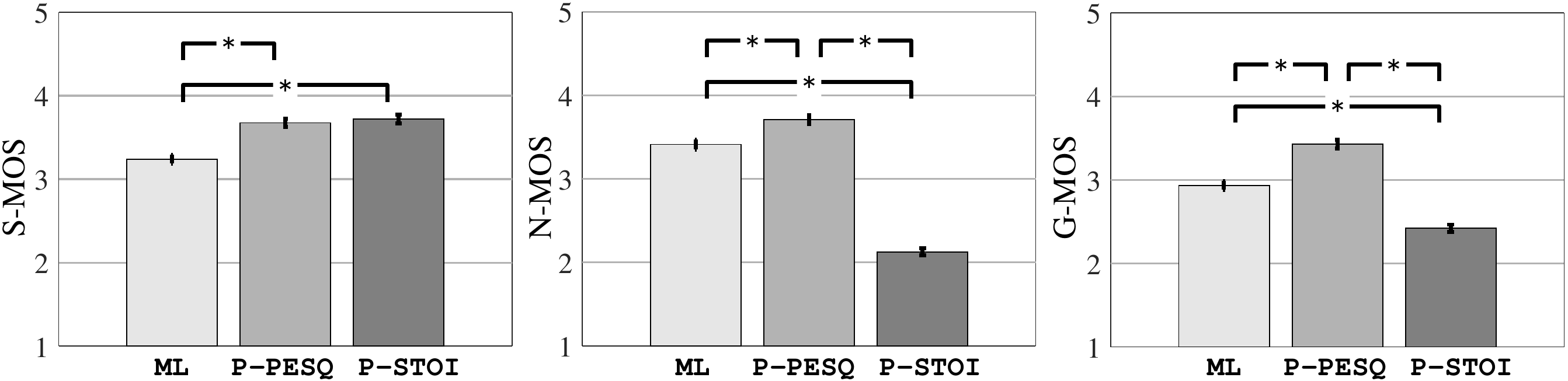}
\caption{Evaluation results of sound-quality test according to ITU-T P.835. 
Bar graphs and error bar indicate average and standard error, respectively.
Asterisks indicate significant difference observed in paired one-sided t-test. 
}
\label{fig:ITU_P835}
\end{figure*}

\begin{figure}[tttt]
\centering
\includegraphics[width=65mm]{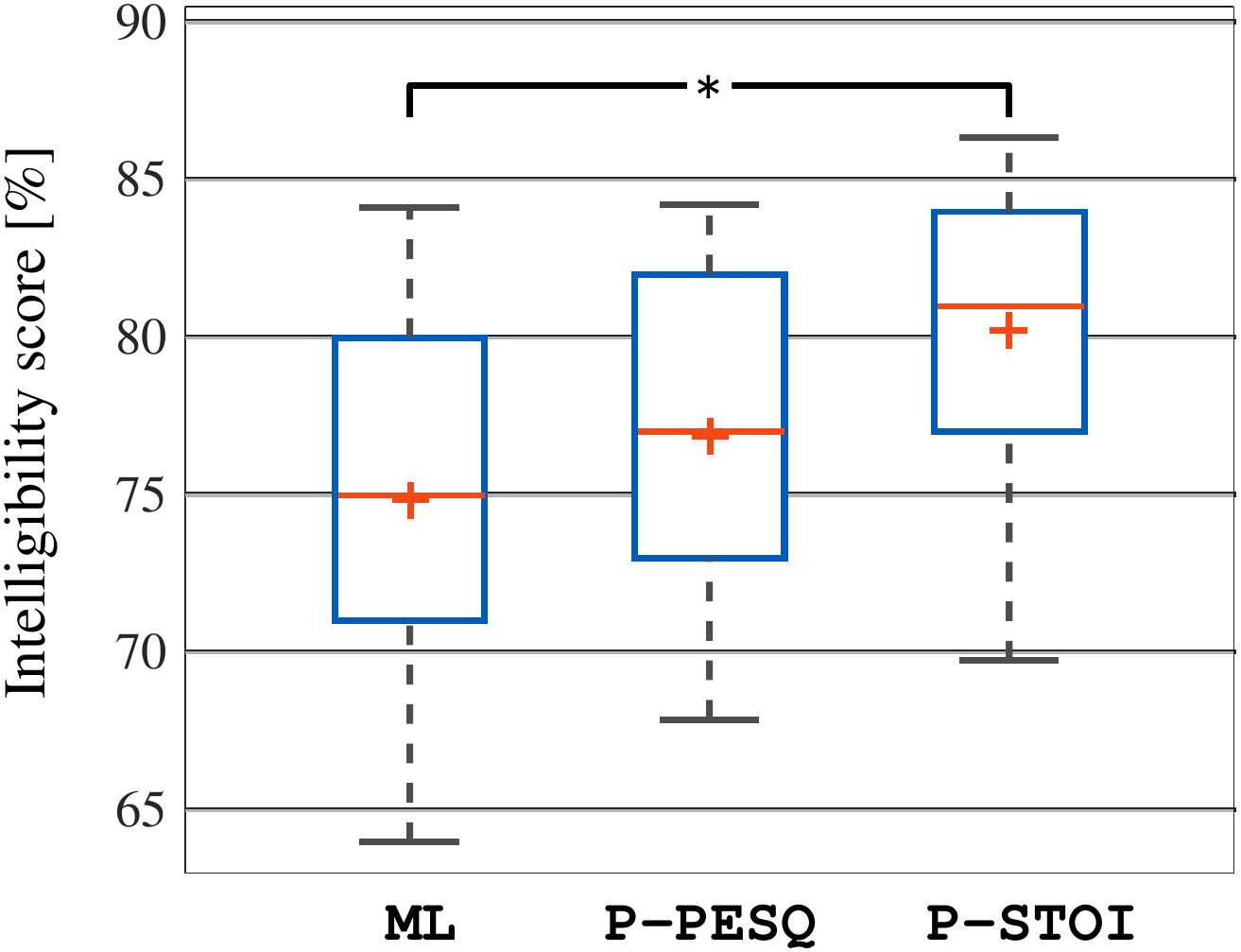}
\caption{
Evaluation results of word-intelligibility test. 
Asterisks indicate significant difference observed in unpaired one-sided t-test. 
}
\label{fig:IntTest}
\end{figure}

Table \ref{tbl:obj_result_sdr} lists the evaluation results of each objective measurement on four noise types and four input SNR conditions. 
The asterisk indicates that the score was significantly higher than both {\tt MMSE} and {\tt ML} in a paired one-sided t-test ($\alpha = 0.05$). 
The SDRs tended to be higher when using the conventional MMSE/ML-based objective function than the proposed method under low SNR conditions.
The PESQ and STOI of {\tt P-PESQ} and {\tt P-STOI} were higher than those of {\tt MMSE} and {\tt ML}, respectively. 
For each method, the PESQ and STOI improved by around 0.1 and 2--5 \%, respectively, and significant differences were observed for all noise and SNR conditions.
These results suggest that the proposed method was able to train the DNN source-enhancement function to directly increase black-box OSQA scores.

In mixed-OSQA experiments, both PESQ and STOI of {\tt P-MIX} were higher than those of {\tt MMSE} and {\tt ML} under almost all noise and SNR conditions.
In the comparison to the results of the mixed-OSQA and single-OSQA ({\it i.e.} {\tt P-PESQ} and {\tt P-STOI}), 
{\tt P-MIX} achieved almost the same or slightly lower PESQ and STOI scores than {\tt P-PESQ} and {\tt P-STOI}, respectively.
In addition, {\tt P-MIX} outperformed STOI and PESQ scores than {\tt P-PESQ} and {\tt P-STOI}, respectively.
These results suggest that the use of the mixed-OSQA would be an effective way to increase multiple-perceptual qualities.

In Table \ref{tbl:obj_result_sdr} we also show that 
the proposed method outperformed the T-F mask selection-based methods \cite{Koizumi_ICASSP_2017} in terms of the target OSQA under almost all noise types and SNR conditions.
Such favorable experimental results would have been observed because of the flexibility of the T-F mask estimation achieved by the proposed method.
In this experiment, the number of the T-F mask template ($=128$) was larger than that used in the previous work ($=32$) \cite{Koizumi_ICASSP_2017}. 
However, since the T-F masks were generated by a combination of the finite number of templates, the patterns of the T-F mask were still limited.
These results suggested that by adopting the policy-gradient method to optimize the parameters of a continuous PDF of the T-F mask processing, the flexibility of the T-F mask estimation was improved.

Figure \ref{fig:result_example} shows examples of the estimated T-F masks and output signal, and Table \ref{tbl:example} lists its objective scores. 
The SNR of the observed signal was adjusted to 0 dB using {\it amusement parks} noise. 
Figure \ref{fig:result_example} shows that the estimated T-F masks reflect the characteristics of each objective function. 
In comparison to the results of {\tt MMSE} and {\tt ML} that reduced the distortion of the target source on average, the T-F mask estimated by {\tt P-PESQ} strongly reduced the residual noise, even when it distorted the target sound at a middle/high frequency ({\it e.g.} Fig. \ref{fig:result_example} white dotted box), and achieved the best PESQ.
In contrast, the T-F mask estimated by {\tt P-STOI} weakly reduced noise to avoid distorting the target source, even when the noise remained in the non-speech frames ({\it e.g.} Fig. \ref{fig:result_example} white dotted circle), and achieved the best STOI. 
This may be because the residual noise degrades the sound quality and the distortion of the target source degrades speech intelligibility. 
The T-F mask estimated by {\tt P-MIX} involved both characteristics and relaxed the disadvantage of {\tt P-PESQ} and {\tt P-STOI}, and both OSQA scores were higher than those of {\tt ML} and {\tt MMSE}.
Namely, speech distortion at a middle/high frequency was reduced ({\it e.g.} Fig. \ref{fig:result_example} white dotted box) and residual noise in the non-speech frames were reduced ({\it e.g.} Fig. \ref{fig:result_example} white dotted circle).

\subsection{Subjective evaluation}
\label{sec:eval_subj}

\subsubsection{Sound quality evaluation}

To investigate the sound quality of the output signals, subjective speech-quality tests were conducted according to ITU-T P.835 \cite{P835}.
In the tests, the participants rated three different factors in the samples:
\begin{itemize}
 \item Speech mean-opinion-score (S-MOS): the speech sample was rated 
 5--not distorted, 
 4--slightly distorted,
 3--somewhat distorted,
 2--fairly distorted, or
 1--very distorted.
 \item Subjective noise MOS (N-MOS): the background of the sample was 
 5--not noticeable, 
 4--slightly noticeable,
 3--noticeable but not intrusive,
 2--somewhat intrusive, or
 1--very intrusive.
 \item Overall MOS (G-MOS): the sound quality of the sample was
 5--excellent, 
 4--good,
 3--fair,
 2--poor, or
 1--bad.
\end{itemize}
Sixteen participants evaluated the sound quality of the output signals of {\tt ML}, {\tt P-PESQ}, and {\tt P-STOI}.
The participants evaluated 20 files for each method; the 20 files consisted of five randomly selected files from the test dataset for each of the four types of noise. 
The input SNR was 6 dB.

Figure \ref{fig:ITU_P835} shows the results of the subjective tests.
For all factors, {\tt P-PESQ} achieved a higher score than {\tt ML}, and statistically significant differences from {\tt ML} were observed in a paired one-sided $t$-test ($p$-value $=0.05$).
The reason for this result suggested that participants may have perceived the degrade of the speech quality from both the speech distortion and the residual noise in speech frame in the output signal of {\tt ML}.
In addition, although there was no statistically significant difference between {\tt P-PESQ} and {\tt P-STOI} in terms of S-MOS score, N-MOS score of {\tt P-STOI} was significantly lower than that of {\tt P-PESQ}.
Thus, G-MOS score of {\tt P-STOI} was also lower than that of {\tt P-PESQ}.
It would be because {\tt P-STOI} weakly reduced noise to avoid distorting the target source, even when the noise remained in the non-speech frames as shown in Sec. IV.C.

\vspace{5pt}
\subsubsection{Speech intelligibility test}

We conducted a word-intelligibility test to investigate speech intelligibility.
We selected 50 low familiarity words from familiarity-controlled word lists 2003 (FW03) \cite{Amano2009} as the test dataset of speech.
The selected dataset consisted of Japanese four-mora words whose accent type was Low-High-High-High.
The noisy test dataset was created by adding a randomly selected noise at SNR of 6 dB from the noisy dataset, which was used in the objective evaluation.
Sixteen participants attempted to write a phonetic transcription for output signals of {\tt ML}, {\tt P-PESQ}, and {\tt P-STOI}.
The percentage of correct answers was used as the intelligibility score.

Figure \ref{fig:IntTest} shows the intelligibility score of each method.
{\tt P-STOI} achieved the highest score.
In addition, statistically significant differences from {\tt ML} were observed in an unpaired one-sided $t$-test ($p$-value $=0.05$).
From both sound-quality and speech-intelligibility tests, we found that the proposed method could improve the specific hearing quality corresponding to the OSQA score used as the objective function.

\section{CONCLUSIONS}
\label{sec:conclusion}

We proposed a training method for the DNN-based source-enhancement function to increase OSQA scores such as the PESQ. 
The difficulty is that the gradient of OSQA scores may not be analytically calculated by simply applying the back-propagation algorithm because most OSQA scores are black boxes. 
To calculate the gradient of the OSQA-based objective function, we formulated a DNN-optimization scheme on the basis of the policy-gradient method. 
In the experiment, 
1) it was revealed that the DNN-based source-enhancement function can be trained using the gradient of the OSQA obtained with the policy-gradient method.
In addition, 
2) the OSQA score and specific hearing quality corresponding to the OSQA score used as the objective function improved.
Therefore, it can be concluded that this method made it possible to use not only analytical objective functions but also black-box functions for the training of the DNN-based source-enhancement function.

Although we focused on maximization of OSQA in this study, the proposed method potentially increases other black-box measurements.
In the future, we will aim to adopt the proposed method to increase other black-box objective measures such as 
the subjective score obtained from a ``human-in-the-loop'' audio-system \cite{MusicAI} and word accuracy of a black-box automatic-speech-recognition system \cite{Watanabe_2014}.
We found that both the PESQ and STOI could increase simultaneously by mixing multiple OSQA scores as an objective function.
In the future, we will also investigate the optimality of the OSQA score and its mixing ratio for the proposed method.

\appendix
\subsection{Deviation of (\ref{eq:total_obj})}
\label{sec:app_A}
We describe the deviation of (\ref{eq:total_obj}). First, as with (\ref{eq:J_Eform}) and (\ref{eq:J_base}), the objective function is defined as the expectation of $\mathcal{B}(\hat{\bm{\mathrm{S}}},\bm{\mathrm{X}})$ as 
\begin{align}
\mathcal{J}(\Theta)	
&= \mathbb{E}_{\hat{\bm{\mathrm{S}}}, \bm{\mathrm{X}}} \left[ \mathcal{B}(\hat{\bm{\mathrm{S}}},\bm{\mathrm{X}}) \right],\\
&= \int p(\bm{\mathrm{X}}) \int \mathcal{B}(\hat{\bm{\mathrm{S}}},\bm{\mathrm{X}}) p(\hat{\bm{\mathrm{S}}}| \bm{\mathrm{X}}, \Theta ) d\hat{\bm{\mathrm{S}}} d \bm{\mathrm{X}} . \label{eq:Utt_J_base}
\end{align}
Then, the gradient of (\ref{eq:Utt_J_base}) can be calculated using a log-derivative trick as
\begin{align}
\nabla _{\Theta} \mathcal{J}(\Theta)	
&= \mathbb{E}_{\bm{\mathrm{X}}} \left[ \mathbb{E}_{\hat{\bm{\mathrm{S}}}| \bm{\mathrm{X}}} \left[ 
\mathcal{B}(\hat{\bm{\mathrm{S}}},\bm{\mathrm{X}}) \nabla _{\Theta} \ln p(\hat{\bm{\mathrm{S}}}| \bm{\mathrm{X}}, \Theta ) \right] \right] \label{eq:Utt_PG_exp}.
\end{align}
By approximating the expectation on $\bm{\mathrm{X}}$ by the average on $\mathcal{I}$ utterances and that of $\hat{\bm{\mathrm{S}}}$ by the average on $K$ times sampling, (\ref{eq:Utt_PG_exp}) can be calculated as 
\begin{align}
\nabla _{\Theta} \mathcal{J}(\Theta)
&\approx 
 \frac{1}{\mathcal{I}} \sum_{\tau=1}^{\mathcal{I}} 
\frac{1}{K} \sum_{k=1}^{K} 
\mathcal{B} ( \hat{\bm{\mathrm{S}}}^{(i, k)}, \bm{\mathrm{X}}^{(i)} )
\nabla _{\Theta} \ln p(\hat{\bm{\mathrm{S}}}^{(i, k)}| \bm{\mathrm{X}}^{(i)}, \Theta ) \label{eq:Utt_nab_J}.
\end{align}
We assume that the output signal on each time frame is calculated independently. Then, $\ln p(\hat{\bm{\mathrm{S}}}| \bm{\mathrm{X}}, \Theta )$ can be reformed to
\begin{equation}
\ln p(\hat{\bm{\mathrm{S}}}| \bm{\mathrm{X}}, \Theta ) = \sum_{\tau = 1}^{T} \ln p(\hat{\bm{S}}_{\tau} | \bm{X}_{\tau}, \Theta ),
\end{equation}
and its gradient can be calculated by
\begin{align}
\nabla _{\Theta} \ln p \left( \hat{\bm{\mathrm{S}}}^{(i, k)}| \bm{\mathrm{X}}^{(i)}, \Theta \right)
&= \sum_{\tau = 1}^{T^{(i)}} \nabla _{\Theta} \ln p(\hat{\bm{S}}^{(i, k)}_{\tau} | \bm{X}^{(i)}_{\tau}, \Theta ), \\
&\approx \frac{1}{T^{(i)}} \sum_{\tau = 1}^{T^{(i)}} \nabla _{\Theta} \ln p(\hat{\bm{S}}^{(i, k)}_{\tau} | \bm{X}^{(i)}_{\tau}, \Theta ). \label{eq:Utt_Nab_ln_q}
\end{align}
To normalize the difference in frame length $T^{(i)}$, we multiplied $1/T^{(i)}$ by the original gradient. 
The log-likelihood function 
$ \ln p(\hat{\bm{S}}^{(i, k)}_{\tau} | \bm{X}^{(i)}_{\tau}, \Theta ) $
can be expanded as
\begin{align}
\ln p(\hat{\bm{S}}^{(i, k)}_{\tau} | \bm{X}^{(i)}_{\tau}, \Theta )
&\mathop{=}^{c} -
\sum_{\omega = 1}^{\Omega} \ln (\sigma_{\omega, \tau}^{2})^{(i)} + 
\frac{ \mathcal{L}_{\Re, \omega, \tau} ^{(i,k)} + \mathcal{L}_{\Im, \omega, \tau} ^{(i,k)} }{2 (\sigma_{\omega, \tau}^{2})^{(i)}},\\
\mathcal{L}_{\Re, \omega, \tau} ^{(i,k)} &=
\left(
\hat{G}_{\omega, \tau}^{(i,k)} \Re\left( X_{\omega, \tau}^{(i)} \right)  - \hat{G}_{\omega, \tau}^{(i)} \Re\left( X_{\omega, \tau}^{(i)} \right) 
\right) ^2,\\
\mathcal{L}_{\Im, \omega, \tau} ^{(i,k)} &=
\left(
\hat{G}_{\omega, \tau}^{(i,k)} \Im \left( X_{\omega, \tau}^{(i)} \right) - \hat{G}_{\omega, \tau}^{(i)} \Im \left( X_{\omega, \tau}^{(i)} \right) 
\right) ^2,
\end{align}
where $\hat{G}_{\omega, \tau}^{(i)}$ and $(\sigma_{\omega, \tau}^{2})^{(i)}$ can be estimated by forward-propagation of the DNN as (\ref{eq:DNN_Gmean})--(\ref{eq:z_return}), and $\hat{G}_{\omega, \tau}^{(i,k)}$ is given by the sampling algorithm of the proposed method.
By using above procedure, $\nabla _{\Theta} \mathcal{J}(\Theta) $ can be calculated by simply applying back-propagation with respect to $\hat{G}_{\omega, \tau}^{(i)}$ and $(\sigma_{\omega, \tau}^{2})^{(i)}$.
Please note that since the simulated output signal $\hat{\bm{S}}^{(i, k)}_{\tau} $ deals with the ``label data'', the back-propagation algorithm is not applied for $\hat{G}_{\omega, \tau}^{(i,k)}$.

\begin{IEEEbiography}
[{\includegraphics[width=1in,height=1.25in,clip,keepaspectratio]{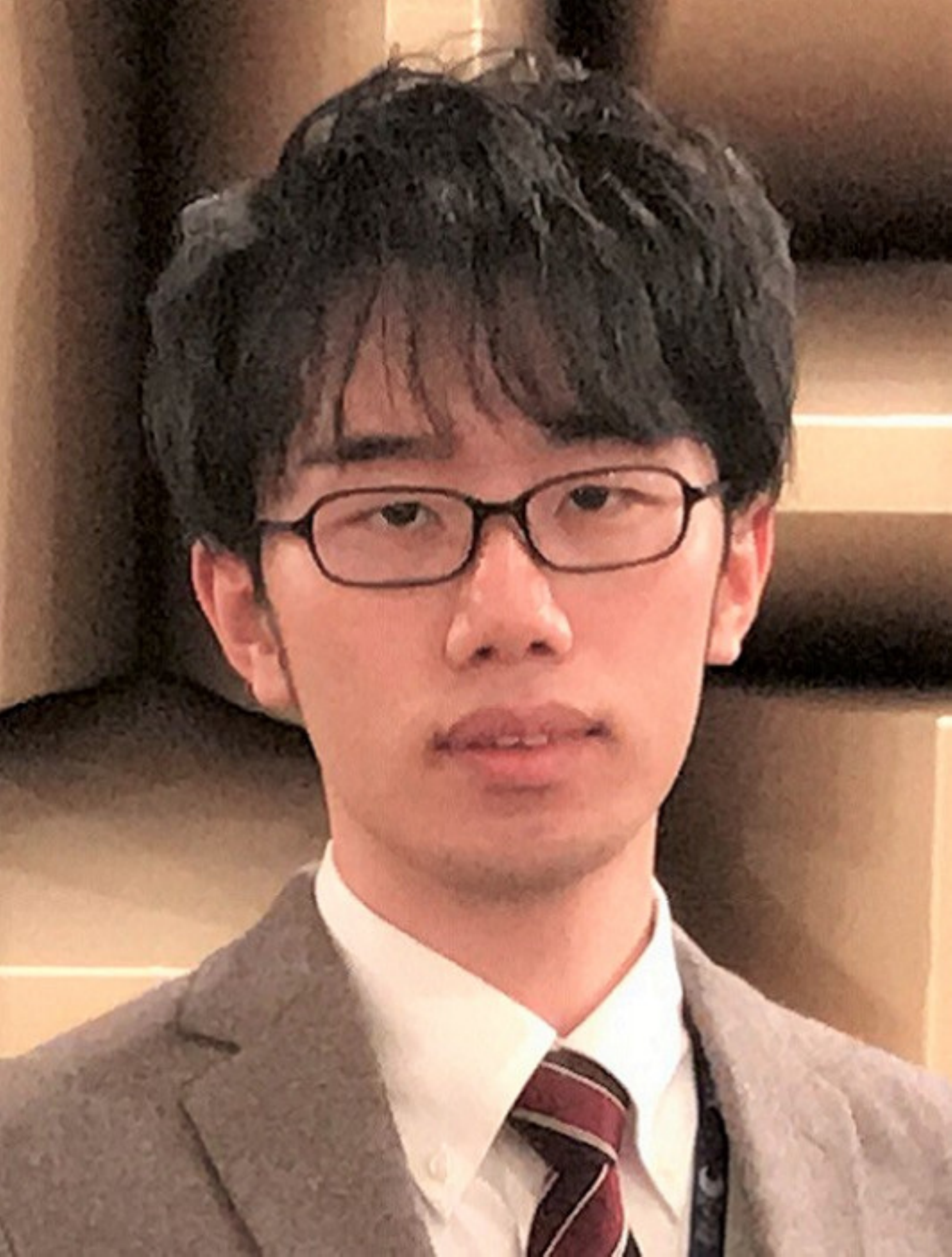}}]
{Yuma Koizumi}(M'15) received the B.S. and M.S. from Hosei University, Tokyo, in 2012 and 2014, and the Ph.D. degree from the University of Electro-Communications in 2017.
Since joining the Nippon Telegraph and Telephone Corporation (NTT) in 2014, he has been researching acoustic signal processing and machine learning. 
He was awarded the IPSJ Yamashita SIG Research Award from the Information Processing Society of Japan (IPSJ) in 2014 and the Awaya Prize from the Acoustical Society of Japan (ASJ) in 2017. 
He is a member of the ASJ and the Institute of Electronics, Information and Communication Engineers (IEICE).
\end{IEEEbiography}

\begin{IEEEbiography}
[{\includegraphics[width=1in,height=1.25in,clip,keepaspectratio]{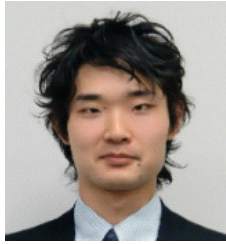}}]
{Kenta Niwa}(M'09)
received his B.E., M.E., and Ph.D. in information science from Nagoya University in 2006, 2008, and 2014. 
Since joining the NTT in 2008, he has been engaged in research on microphone array signal processing as a research engineer at NTT Media Intelligence Laboratories. 
From 2017, he is also a visiting researcher at Victoria University of Wellington, New Zealand. 
He was awarded the Awaya Prize by the ASJ in 2010. 
He is a member of the ASJ and the IEICE. 
\end{IEEEbiography}

\begin{IEEEbiography}
[{\includegraphics[width=1in,height=1.25in,clip,keepaspectratio]{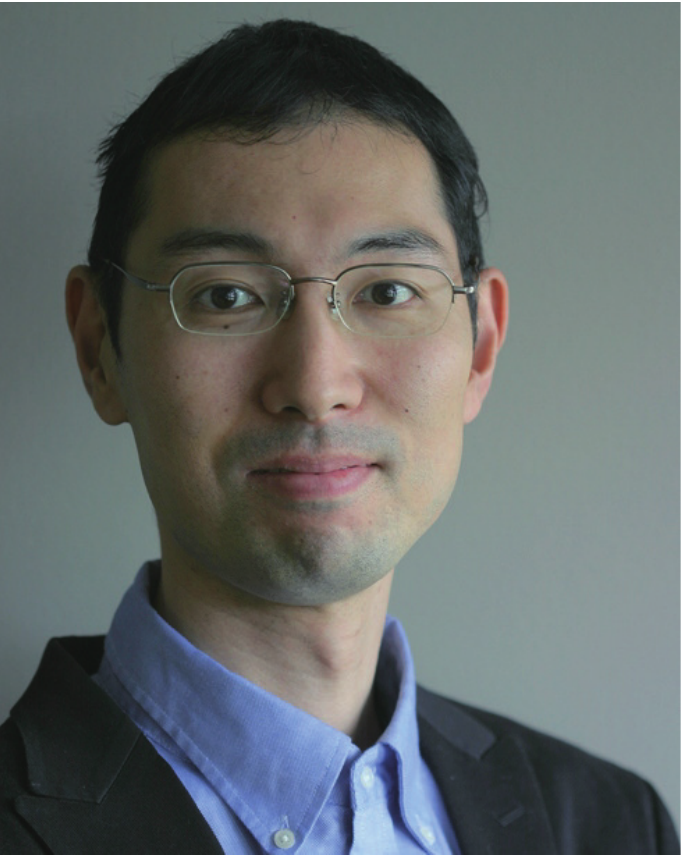}}]
{Yusuke Hioka} (S'04-M'05-SM'12) received his B.E., M.E., and Ph.D. degrees in engineering in 2000, 2002, and 2005 from Keio University, Yokohama, Japan. 
From 2005 to 2012, he was with the NTT Cyber Space Laboratories (now NTT Media Intelligence Laboratories), NTT in Tokyo. 
From 2010 to 2011, he was also a visiting researcher at Victoria University of Wellington, New Zealand. 
In 2013 he permanently moved to New Zealand and was appointed as a Lecturer at the University of Canterbury, Christchurch. 
Then in 2014, he joined the Department of Mechanical Engineering, the University of Auckland, Auckland, where he is currently a Senior Lecturer. 
His research interests include audio and acoustic signal processing especially microphone arrays, room acoustics, human auditory perception and psychoacoustics. 
He is a Senior Member of IEEE and a Member of the Acoustical Society of New Zealand, ASJ, and the IEICE.
\end{IEEEbiography}

\begin{IEEEbiography}
[{\includegraphics[width=1in,height=1.25in,clip,keepaspectratio]{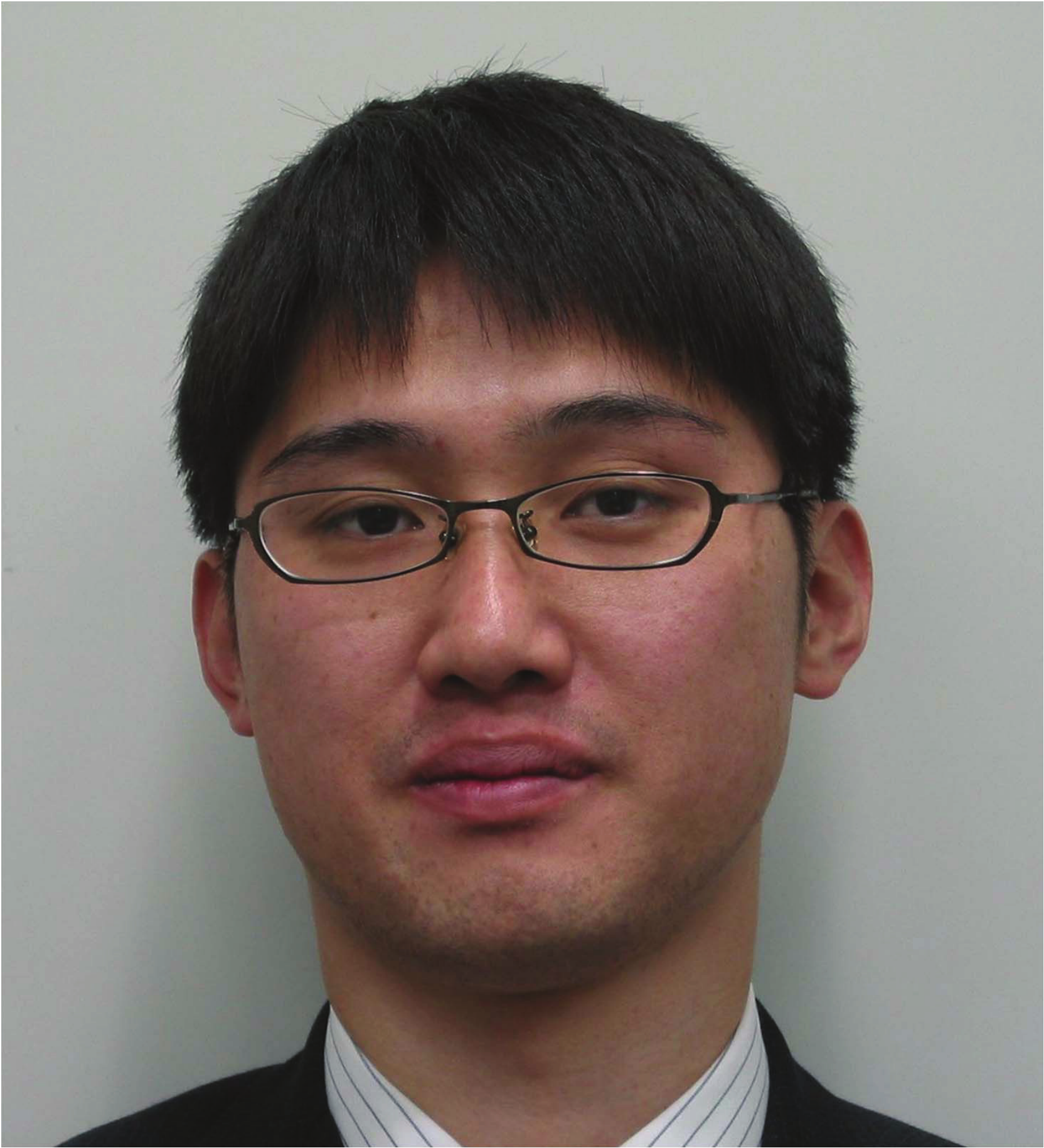}}]
{Kazunori Kobayashi} 
received the B.E., M.E., and Ph.D. degrees in Electrical and Electronic System Engineering from Nagaoka University of Technology in 1997, 1999, and 2003. 
Since joining NTT in 1999, he has been engaged in research on microphone arrays, acoustic echo cancellers and hands-free systems. 
He is now Senior Research Engineer of NTT Media Intelligence Laboratories. 
He is a member of the ASJ and the IEICE.
\end{IEEEbiography}

\begin{IEEEbiography}
[{\includegraphics[width=1in,height=1.25in,clip,keepaspectratio]{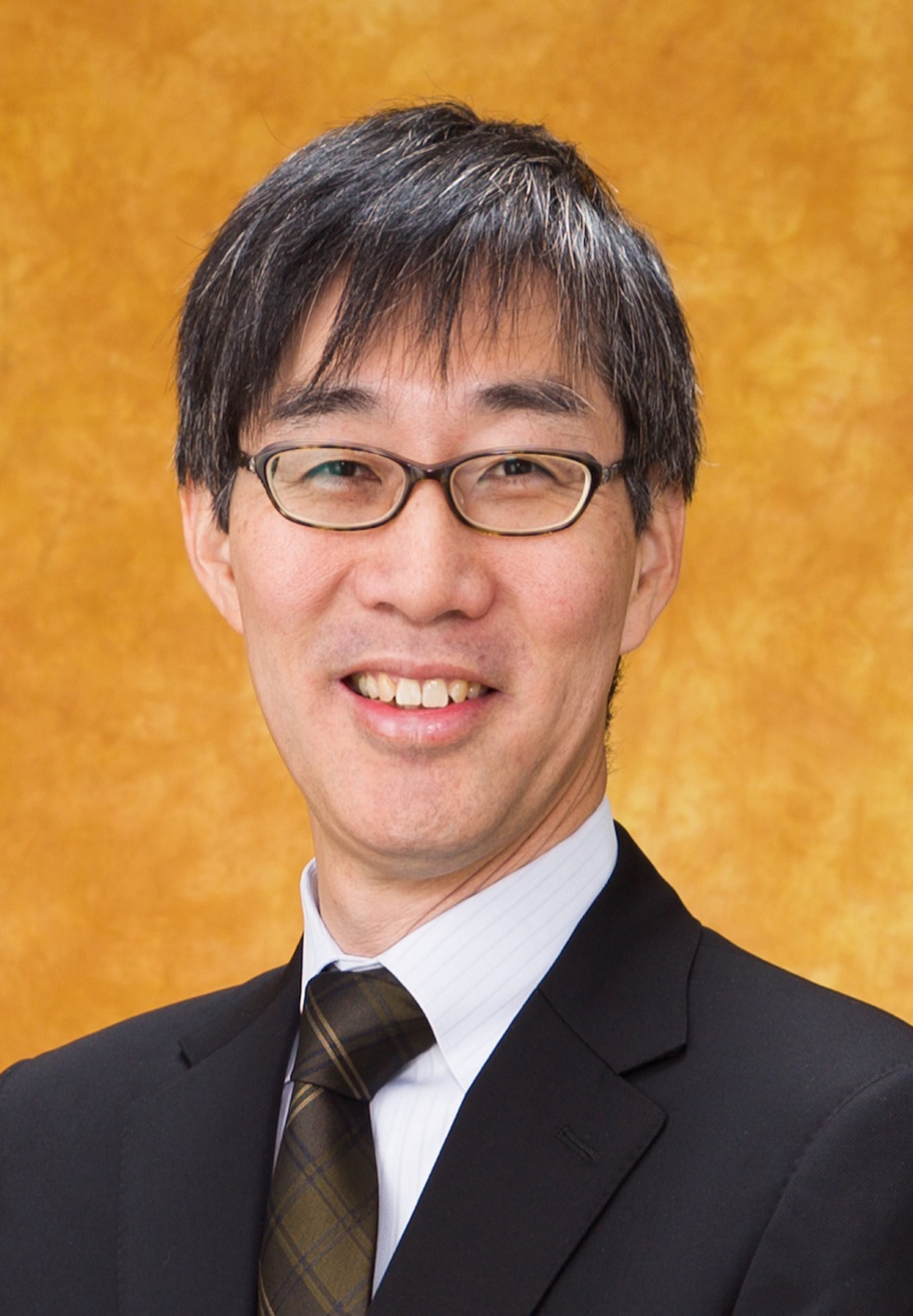}}]
{Yoichi Haneda} (M'97-SM'06) received the B.S., M.S., and Ph.D. degrees from Tohoku University, Sendai, in 1987, 1989, and 1999. From 1989 to 2012, he was with the NTT, Japan. 
In 2012, he joined the University of Electro-Communications, where he is a Professor. 
His research interests include modeling of acoustic transfer functions, microphone arrays, loudspeaker arrays, and acoustic echo cancellers. 
He received paper awards from the ASJ and from the IEICE of Japan in 2002. Dr. Haneda is a senior member of IEICE, and a member of AES, ASA and ASJ.
\end{IEEEbiography}


\begin{thebibliography}{99}


\bibitem{spenh} J.~Benesty, S.~Makino, and J.~Chen, Eds.,
``Speech enhancement,''
Springer, 2005.
\bibitem{Ephraim_1984}
Y.~Ephraim and D.~Malah,
``Speech enhancement using a minimum mean-square error short-time spectral amplitude estimator,''
{\it IEEE Trans. Audio, Speech and Language Processing}, pp.1109--1121, 1984.
\bibitem{zelinski88} R.~Zelinski
``A microphone array with adaptive post-filtering for noise reduction in reverberant rooms,''
in {\it Proc. ICASSP}, pp. 2578 --2581, 1988.
\bibitem{Hioka2013} Y.~Hioka, K.~Furuya, K.~Kobayashi, K.~Niwa and Y.~Haneda,
``Underdetermined sound source separation using power spectrum density estimated by combination of directivity gain,''
{\it IEEE Trans. Audio, Speech and Language Processing}, pp.1240--1250, 2013.
\bibitem{Niwa2016} K.~Niwa, Y.~Hioka, and K.~Kobayashi,
``Optimal Microphone Array Observation for Clear Recording of Distant Sound Sources,''
{\it IEEE/ACM Trans. Audio, Speech and Language Processing}, pp.1785--1795, 2016.
\bibitem{Lightburn_2017} L.~Lightburn, E.~D.~Sena, A.~Moore, P.~A.~Naylor, M.~Brookes,
``Improving the perceptual quality of ideal binary masked speech,''
in {\it Proc. ICASSP}, 2017.
\bibitem{Yoshioka2012} T.~Yoshioka, A.~Sehr, M.~Delcroix, K.~Kinoshita, R.~Maas, T.~Nakatani, and W.~Kellermann,
``Making machines understand us in reverberant rooms: robustness against reverberation for automatic speech recognition,''
{\it IEEE Signal Processing Magazine}, pp. 114--126, 2012.
\bibitem{Narayanan2013} A.~Narayanan and D.~Wang,
``Ideal ratio mask estimation using deep neural networks for robust speech recognition,''
in {\it Proc. ICASSP}, 2013.
\bibitem{Ochiai_2017}
T.~Ochiai, S.~Watanabe, T.~Hori, and J.~R.~Hershey,
``Multichannel End-to-end Speech Recognition,''
in {\it Proc. ICML}, 2017.


\bibitem{Kobayashi_08} K.~Kobayashi, Y.~Haneda, K.~Furuya, and A.~Kataoka,
``A hands-free unit with noise reduction by using adaptive beamformer,''
{\it IEEE Trans. on Consumer Electronics}, Vol.54-1, 2008.
\bibitem{Hioka_12} Y.~Hioka, K.~Furuya, K.~Kobayashi, S.~Sakauchi, and Y.~Haneda,
``Angular region-wise speech enhancement for hands-free speakerphone,''
{\it IEEE Trans. on Consumer Electronics}, Vol.58-4, 2012.
\bibitem{Moore_2003} B.~C.~J.~Moore,
``Speech processing for the hearing-impaired: successes, failures, and implications for speech mechanisms,''
{\it Speech Communication}, Vol. 41, Issue 1, pp.81--91, 2003.
\bibitem{Wang_2008} D.~L.~Wang, 
``Time-frequency masking for speech separation and its potential for hearing aid design,''
{\it Trends in Amplification}, vol. 12, pp. 332--353, 2008.
\bibitem{Tao_2016} T.~Zhang, F.~Mustiere, and C.~Micheyl,
``Intelligent Hearing Aids: The Next Revolution,''
In {\it Proc. EMBC,} 2016.
\bibitem{Zhao_2016} Y.~Zhao, D.~Wang, I.~Merks, and T.~Zhang,
``DNN-based enhancement of noisy and reverberant speech,''
In {\it Proc. ICASSP}, 2016.
\bibitem{obj_base} R. Oldfield, B. Shirley and J. Spille,
``Object-based audio for interactive football broadcast,''
{\it Multimedia Tools and Applications}, Vol. 74, pp.2717--2741, 2015.
\bibitem{Koizumi_2017}
Y. Koizumi, K. Niwa, Y. Hioka, K. Kobayashi and H. Ohmuro,
``Informative acoustic feature selection to maximize mutual information for collecting target sources,''
{\it IEEE/ACM Trans. Audio, Speech and Language Processing}, pp.768--779, 2017.


\bibitem{LeCun_DL_2015} Y.~LeCun, Y.~Bengio, and G.~Hinton,
``Deep Learning,''
{\it Nature}, 521, pp.436--444, 2015.

\bibitem{Weninger_2015} 
F.~Weninger, H.~Erdogan, S.~Watanabe, E.~Vincent, J.~L.~Roux, J.~R.~Hershey, and B.~Schuller,
``Speech Enhancement with LSTM Recurrent Neural Networks and its Application to Noise-Robust ASR,''
in {\it Proc. LVA/ICA}, 2015.
\bibitem{Erdogan_2015} 
H.~Erdogan, J.~R.~Hershey, S.~Watanabe, and J.~L.~Roux,
``Phase-sensitive and recognition-boosted speech separation using deep recurrent neural networks,'' 
in {\it Proc. ICASSP}, 2015.
\bibitem{Will_cIRM_2017} D.~S.~Williamson and D.~L.~Wang,
``Time-frequency masking in the complex domain for speech dereverberation and denoising,''
{\it IEEE/ACM Trans. Audio, Speech and Language Processing}, 2017.
\bibitem{Zhao_2018} Y.~Zhao, B.~Xu, R.~Giri, and T.~Zhang,
``Perceptually Guided Speech Enhancement using deep neural networks,''
in {\it Proc. ICASSP}, 2018.

\bibitem{Xu_2014} Y.~Xu, J.~Du, L.~R.~Dai, and C.~H.~Lee, 
``An experimental study on speech enhancement based on deep neural networks,''
{\it IEEE Signal Processing Letters}, pp.65--68, 2014.
\bibitem{Y_xu_2015_DNN}
Y.~Xu, J.~Du, L.~R.~Dai and C.~H.~Lee,
``A regression approach to speech enhancement based on deep neural networks,''
{\it IEEE/ACM Trans. Audio, Speech and Language Processing}, pp.7--19, 2015.
\bibitem{Xu_2015} Y.~Xu, J.~Du, Z.~Huang, L.~R.~Dai, and C.~H.~Lee,
``Multi-objective learning and mask-based post-processing for deep neural network based speech enhancement,''
in {\it Proc. INTERSPEECH}, 2015.
\bibitem{Gao_2016} T.~Gao, J.~Du, L.~R.~Dai, and C.~H.~Lee, 
``SNR-Based Progressive Learning of Deep Neural Network for Speech Enhancement,''
in {\it Proc. INTERSPEECH}, 2016.
\bibitem{QWang_2018} Q.~Wang, J.~Du, L.~R.~Dai and C.~H.~Lee, 
``A multiobjective learning and ensembling approach to high-performance speech enhancement with compact neural network architectures,''
{\it IEEE/ACM Trans. Audio, Speech and Language Processing}, pp.1181--1193, 2018.
\bibitem{Kawase2016}
T.~Kawase, K.~Niwa, K.~Kobayashi, and Y.~Hioka,
``Application of neural network to source PSD estimation for Wiener filter based sound source separation,''
in {\it Proc. IWAENC}, 2016.
\bibitem{DNN_Niwa_2017} K.~Niwa, Y.~Koizumi, T.~Kawase, K.~Kobayashi and Y.~Hioka,
``Supervised Source Enhancement Composed of Non-negative Auto-Encoders and Complementarity Subtraction''
in {\it Proc. ICASSP}, 2017.
\bibitem{Smagradis_2017} P.~Smaragdis and S.~Venkataramani,
``A Neural Network Alternative to Non-Negative Audio Models,''
in {\it Proc. ICASSP}, 2017.
\bibitem{Chai_2017} L.~Chai, J.~Du and Y.~Wang,
``Gaussian Density Guided Deep Neural Network For Single-Channel Speech Enhancement,''
in {\it Proc. MLSP}, 2017.


\bibitem{Kinoshita_2017} K.~Kinoshita, M.~Delcroix, A.~Ogawa, T.~Higuchi, and T.~Nakatani,
``Deep Mixture Density Network for Statistical Model-based Feature Enhancement,''
in {\it Proc. ICASSP}, 2017.
\bibitem{Arie_2016} A.~A.~Nugraha, A.~Liutkus, and E.~Vincent,
``Multichannel Audio Source Separation With Deep Neural Networks,''
{\it IEEE/ACM Trans. Audio, Speech and Language Processing}, 2016.
\bibitem{Hershey} J.~Hershy, Z.~Chen, J.~L.~Roux, and S.~Watanabe,
``Deep clustering: Discriminative embeddings for segmentation and separation,''
In {\it Proc. ICASSP}, 2016.
\bibitem{Pascual_2017} S.~Pascual, A.~Bonafonte, and J.~Serra,
``SEGAN: Speech Enhancement Generative Adversarial Network,''
In {\it Proc INTERSPEECH}, 2017.







\bibitem{back_prop} D.~E.~Rumelhart, G.~E.~Hinton, E.~Geoffrey and R.~J.~Williams,
``Learning representations by back-propagating errors,''
{\it Nature}, 323, pp.533--536, 1986.
\bibitem{PESQ} ITU-T Recommendation P.862,
``Perceptual evaluation of speech quality (PESQ): An objective method for end-to-end speech quality assessment of narrow-band telephone networks and speech codecs,'' 2001.
\bibitem{STOI} C.~H.~Taal, R.~C.~Hendriks, R.~Heusdens, and J.~Jensen,
``An Algorithm for Intelligibility Prediction of Time-Frequency Weighted Noisy Speech,''
{\it IEEE Transactions on Audio, Speech and Language Processing}, Vol. 19, pp.2125--2136, 2011.
\bibitem{Koizumi_ICASSP_2017} Y.~Koizumi, K.~Niwa, Y.~Hioka, K.~Kobayashi and Y.~Haneda,
`DNN-based Source Enhancement Self-optimized by Reinforcement Learning using Sound Quality Measurements,''
in {\it Proc. ICASSP}, 2017.
\bibitem{RL_text} E.~S.~Sutton and A.~G.~Barto,
``Reinforcement Learning: An Introduction,''
{\it A Bradford Book,} 1998.
\bibitem{alpha_go} 
D.~Silver, A.~Huang, C.~J.~Maddison, A.~Guez, L.~Sifre, G.~Driessche, J.~Schrittwieser, I.~Antonoglou, V.~Panneershelvam, M.~Lanctot, S.~Dieleman, D.~Grewe, J.~Nham, N.~Kalchbrenner, 
I.~Sutskever, T.~Lillicrap, M.~Leach, K.~Kavukcuoglu, T.~Graepel and D.~Hassabis,
`` Mastering the game of Go with deep neural networks and tree search,''
{\it Nature}, pp.484---489, 2016.

\bibitem{Policy_Grad} R.~J.~Williams, 
``Simple Statistical Gradient-Following Algorithms for Connectionist Reinforcement Learning,''
{\it Machine Learning}, Vol. 8, 1992.
\bibitem{advantage_AC} R.~S.~Sutton, D.~McAllester, S.~Singh, and Y.~Mansour,
``Policy Gradient Methods for Reinforcement Learning with Function Approximation,''
In {\it Proc. NIPS}, 1999.
\bibitem{Mersenne} M.~Matsumoto and T.~Nishimura, 
``Mersenne Twister: A 623-dimensionally Equidistributed Uniform Pseudorandom Number Generator,''
ACM Trans. on Modeling and Computer Simulations, 1998.




\bibitem{ATR_database} A.~Kurematsu, K.~Takeda, Y.~Sagisaka, S.~Katagiri, H.~ Kuwabara, and K. Shikano,
``ATR Japanese speech database as a tool of speech recognition and synthesis,'' 
{\it Speech communication}, pp.357--363, 1990.
\bibitem{CHiME} J. Barker, R. Marxer, E. Vincent and S. Watanabe,
``The third `CHiME' speech separation and recognition challenge: dataset, task and baseline,''
in {\it Proc. ASRU}, 2015.
\bibitem{adam} D.~P.~Kingma and J.~Ba,
``Adam: A Method for Stochastic Optimization,''
in {\it Proc ICLR}, 2015. 
\bibitem{Weninger_2014} F.~Weninger, J.~R.~Hershey, J.~L.~Roux and B.~Schuller,
``Discriminatively Trained Recurrent Neural Networks for Single-Channel Speech Separation,''
in {\it Proc. GlobalSIP}, 2014.

\bibitem{pre-train} F. Seide, G. Li, X. Chen and D. Yu,
``Feature engineering in context-dependent deep neural networks for conversational speech transcription,''
in {\it Proc. ASRU}, pp. 24--29, 2011.

\bibitem{Miyazaki_2012} R.~Miyazaki, H.~Saruwatari, T.~Inoue, Y.~Takahashi, K.~Shikano and K.~Kondo,
``Musical-Noise-Free Speech Enhancement Based on Optimized Iterative Spectral Subtraction,''
{\it IEEE Transactions on Audio, Speech and Language Processing}, Vol. 20, pp.2080--2094, 2012.
\bibitem{Cohen_2002} I.~Cohen,
``Optimal Speech Enhancement Under Signal Presence Uncertainty Using Log-Spectral Amplitude Estimator,''
{\it IEEE Signal Processing Letters}, Vol. 9, pp.113--116, 2002.
\bibitem{Emmanuel_2010} E.~Vincent,
``An Experimental Evaluation of Wiener Filter Smoothing Techniques Applied to Under-Determined Audio Source Separation,''
in {\it Proc. LVA/ICA}, 2010.
\bibitem{Niwa_IWAENC} K.~Niwa, Y.~Hioka, and K.~Kobayashi,
``Post-Filter Design for Speech Enhancement in Various Noisy Environments,''
in {\it Proc IWAENC}, 2014.

\bibitem{Vincent2006} E.~Vincent, R.~Gribonval and C.~Fevotte,
``Performance measurement in blind audio source separation,''
{\it IEEE Trans. Audio, Speech and Language Processing}, 14(4), pp.1462--1469, 2006.


\bibitem{P835} ITU-T Recommendation P.835,
``Subjective test methodology for evaluating speech communication systems that include noise suppression algorithm,'' 
2003.
\bibitem{Amano2009}
S.~Amano, S.~Sakamoto, T.~Kondo, and Y.~Suzuki,
``Development of familiarity-controlled word lists 2003 (FW03) to assess spoken-word intelligibility in Japanese,''
{\it Speech Communication}, pp. 76--82, 2009.



\bibitem{MusicAI} K.~Niwa, K.~Ohtani and K,~Takeda,
``Music Staging AI,''
in {\it Proc. ICASSP}, 2017.
\bibitem{Watanabe_2014} S.~Watanabe and J.~L.~Roux,
``Black Box Optimization for Automatic Speech Recognition,''
in {\it Proc. ICASSP}, 2014.





\end{thebibliography}
\end{document}